\documentclass[lettersize,journal]{IEEEtran}
\usepackage{amsmath,amsfonts}
\usepackage{algorithm}
\usepackage{algorithmic}
\usepackage{array}
\usepackage[caption=false,font=normalsize,labelfont=sf,textfont=sf]{subfig}
\usepackage{textcomp}
\usepackage{stfloats}
\usepackage{url}
\usepackage{verbatim}
\usepackage{graphicx}
\usepackage{cite}

\usepackage{booktabs}
\usepackage{xcolor}
\usepackage{smile_symbols}
\usepackage{arydshln}

\usepackage{hyperref}
\hypersetup{
    colorlinks=true,
    linkcolor=blue,
    filecolor=magenta,      
    urlcolor=cyan,
}
\usepackage{multirow}

\begin{document}

\title{Unveiling the Unseen: A Comprehensive Survey on Explainable Anomaly Detection in Images and Videos}

\author{Yizhou~Wang$^*$,~\IEEEmembership{Student Member,~IEEE,}
        Dongliang Guo$^*$,
        Sheng Li,~\IEEEmembership{Senior Member,~IEEE,}
        Octavia Camps,~\IEEEmembership{Member,~IEEE,} Yun~Fu,~\IEEEmembership{Fellow,~IEEE}
\IEEEcompsocitemizethanks{
\IEEEcompsocthanksitem $^*$ denotes equal contribution.
\IEEEcompsocthanksitem Y. Wang and O. Camps are with the Department of Electrical and Computer Engineering, Northeastern University, Boston,
MA, 02115. E-mail: wyzjack990122@gmail.com, camps@ece.northeastern.edu.
\IEEEcompsocthanksitem D. Guo and S. Li are with the School of Data Science, University of Virginia, VA. E-mail: 	dongliang.guo@virginia.edu, vga8uf@virginia.edu.%
\IEEEcompsocthanksitem Y. Fu is with the Department of Electrical and Computer Engineering and Khoury College of Computer Science, Northeastern University, Boston,
MA, 02115. Email: yunfu@ece.neu.edu.}}

\markboth{Journal of \LaTeX\ Class Files,~Vol.~14, No.~8, August~2021}%
{Shell \MakeLowercase{\textit{et al.}}: A Sample Article Using IEEEtran.cls for IEEE Journals}


\maketitle

\begin{abstract}
Anomaly detection and localization in visual data, including images and videos, are crucial in machine learning and real-world applications. Despite rapid advancements in visual anomaly detection (VAD), interpreting these often black-box models and explaining why specific instances are flagged as anomalous remains challenging. This paper provides the first comprehensive survey focused specifically on explainable 2D visual anomaly detection (X-VAD), covering methods for both images (IAD) and videos (VAD). We first introduce the background of IAD and VAD. Then, as the core contribution, we present a thorough literature review of explainable methods, categorized by their underlying techniques (e.g., attention-based, generative model-based, reasoning-based, foundation model-based). We analyze the commonalities and differences in applying these methods across image and video modalities, highlighting modality-specific challenges and opportunities for explainability. Additionally, we summarize relevant datasets and evaluation metrics, discussing both standard performance metrics and emerging approaches for assessing explanation quality (e.g., faithfulness, stability). Finally, we discuss promising future directions and open problems, including quantifying explanation quality, explaining diverse AD paradigms (SSL, zero-shot), enhancing context-awareness, leveraging foundation models responsibly, and addressing real-world constraints like efficiency and robustness. A curated collection of related resources is available at \url{https://github.com/wyzjack/Awesome-XAD}.
\end{abstract}

\begin{IEEEkeywords}
Anomaly Detection, Explainable Machine Learning, Image Processing, Video Processing, Computer Vision.
\end{IEEEkeywords}

\section{Introduction}

Anomaly detection plays a crucial role in society. Nowadays, with the development technique of computer vision, many researchers try to address visual anomaly detection tasks, specifically focusing on 2D data which primarily includes images and videos (referred to as 2D visual anomaly detection hereafter). Relevant tasks include image anomaly detection (IAD)~\cite{reiss2021panda,pang2021explainable,roth2022towards,cohen2022transformaly} and video anomaly detection (VAD)~\cite{Georgescu2021ssmtl,Liu_2021_HFVAD,chen2022comprehensive,liu2018future,yu2020cloze,wang2022jigsaw-vad,Georgescu2021Background} (3D point cloud anomaly detection works~\cite{bergmann2021mvtec,horwitz2023back,liu2023real3d,wang2025towards} are out of the scope of this paper). 2D visual anomaly detection aims to identify unusual or abnormal patterns within these visual modalities. It finds applications in various domains, such as surveillance, security, medical imaging, and industrial inspection. The fundamental idea is to train a model on normal patterns and then identify deviations in new, unseen data. Due to the high-dimensional nature of visual data, traditional statistical methods effective for tabular data often fall short. Consequently, deep learning-based methods are extensively employed. Common architectures like Autoencoders (AEs) and Variational Autoencoders (VAEs) learn compact representations or generative models of normal data, enabling the identification of outliers based on reconstruction errors or likelihood scores.

While deep learning models show promise in identifying visual anomalies, their inherent "black-box" nature often limits their explainability. Explainability, in this context, refers to the ability of a model to provide understandable reasons for its decisions to human users~\cite{vilone2018explaining, molnar2019machine}. For 2D visual anomaly detection, an explainable model should offer clear and faithful justifications for why a specific image region or video segment is flagged as anomalous. The lack of transparency hinders trust and practical adoption, particularly in safety-critical domains~\cite{langone2020interpretable,10.1145/3609333}. For instance, in medical diagnosis, simply predicting a disease from an X-ray without explaining \textit{which} features led to the conclusion is insufficient for clinical acceptance. Providing explanations is becoming an ethical and regulatory necessity~\cite{belle2021interpretability}. Furthermore, understanding \textit{why} something is anomalous is often crucial for troubleshooting and remediation in applications like industrial quality control. Therefore, developing explainable 2D visual anomaly detection methods is essential.

To address the need for explainability, various methods have emerged recently. These approaches differ significantly based on the specific task, data modality (image vs. video), and the target audience for the explanation. For example, in surveillance video anomaly detection,~\cite{szymanowicz2022discrete} employed an autoencoder and used the reconstruction error map (saliency map) combined with off-the-shelf object/action recognition models to provide post-hoc explanations. This survey focuses specifically on explainable methods developed for or applied to 2D visual anomaly detection tasks, categorizing them primarily based on the underlying technique used to achieve explainability within the context of IAD and VAD.


Anomaly detection in general is a well-studied field with numerous surveys~\cite{ruff2021deep,mohammadi2021image,ramachandra2020survey,jiang2022survey,cao2024survey}. Explainable AI (XAI) also has comprehensive reviews covering various methods and metrics~\cite{carvalho2020interpretable, molnar2019machine, linardatos2020taxonomy, vilone2018explaining, xie2021survey, belle2021interpretability, tjoa2017interpretability}. However, surveys specifically focusing on \textit{explainable anomaly detection} are fewer~\cite{10.1145/3609333, kauffmann2020towards}. Notably, the recent survey by Li et al.~\cite{li2023survey} provides a valuable overview of explainable anomaly detection across different data types. While insightful, its broad scope means it provides less depth on the unique challenges and techniques specific to the visual domain (images and videos). Our survey aims to fill this gap by concentrating exclusively on explainable \textit{2D visual} anomaly detection. We delve deeper into the specific explainability methods tailored for or applied to images and videos, analyze the commonalities and differences between IAD and VAD explainability, discuss domain-specific datasets and evaluation metrics relevant to explainability, and propose future directions pertinent to visual data. Our literature search aimed for comprehensive coverage of papers explicitly addressing explainability in visual anomaly detection, retrieved through systematic searches of major computer vision and machine learning venues and citation chasing. This focused approach allows for a more detailed characterization of the state-of-the-art in this rapidly evolving subfield, offering a refined taxonomy and actionable insights for researchers and practitioners working with visual data.

In summary, our paper presents several noteworthy contributions summarized as follows:
\begin{itemize}
    \item We provide the first focused survey on explainable methods specifically for image and video anomaly detection.
    \item We propose a taxonomy tailored to these explainable visual methods, comparing them thoroughly, outlining their advantages, disadvantages, and specific applicability to IAD versus VAD.
    \item We analyze the unique aspects of explainability in visual domain, including relevant datasets and evaluation considerations beyond standard anomaly detection metrics.
    \item We identify and discuss potential future research directions specific to explainable 2D visual anomaly detection.
\end{itemize}

The remainder of this survey is structured as follows. We first introduce the background of explainable 2D anomaly detection and some concepts in Sec.~\ref{2-overview}. Then, we summarize the existing methods for explainable 2D anomaly detection in Sec.~\ref{sec: 3-methods}. Next, we conclude the commons and differences of explainable image and video anomaly detection approaches in Sec.~\ref{sec: 4-differences-and-commonalities}. We list the commonly used datasets for 2D visual anomaly detection in Sec.~\ref{sec: 5-datasets} and the standard evaluation metrics in Sec.~\ref{sec: 6-metrics}. Finally, we point out potential directions for this topic in Sec.~\ref{sec: 7-future direction}.
\section{Background}
\label{2-overview}


2D anomaly detection pertains to the process of detecting abnormal patterns in images or videos. 2D anomaly detection approaches are usually trained on normal samples only and concentrate on learning the representation of anomaly-free data and detecting deviations from the learned model as anomalies. Based on the data type, 2D anomaly detection can be generally classified into image anomaly detection (IAD) and video anomaly detection (VAD).

\paragraph{\bf Image anomaly detection} is of considerable significance in both academia and industry, with wide-range application scenarios including industrial inspection, medical health, and website management, etc. Image anomaly detection can be broadly classified into two sub-tasks: semantic anomaly detection~\cite{salehi2021multiresolution,wang2022self,tack2020csi,reiss2021panda} and sensory anomaly detection in images~\cite{bergmann2020uninformed,venkataramanan2020attention,roth2022towards} and some approaches can possess both functions. For one-class classification, classification datasets, such as CIFAR-10 and CIFAR-100~\cite{krizhevsky2009learning}, are usually employed to create a one-class or leave-one-out setting or dataset-wise out-of-distribution setting.~\cite{tack2020csi} put forward a distribution-shifting transformations-based contrastive learning scheme.~\cite{wang2022self} seamlessly combine feature-level self-supervised learning with ad hoc feature adversarial perturbation for state-of-the-art performance. For both image-level anomaly detection and localization, the most commonly used dataset is MVTec AD~\cite{bergmann2019mvtec}.~\cite{bergmann2020uninformed} pioneers image-level anomaly detection and segmentation through the disparity in output between the teacher and student models. More recently,~\cite{roth2022towards} exploits an ImageNet-pretrained network for patch-level feature embedding extraction and proposes an effective core-set selection and representation distance calculation for anomaly score design, exhibiting the state-of-the-art performance so far.
\paragraph{\bf Anomaly detection in videos} (e.g., surveillance videos) is a challenging yet significant task in multiple domains, such as public safety, healthcare, and ecology. The classic setting is that merely normal video clips are provided for training, and during testing the video sequences are composed of both normal and abnormal frames. The VAD algorithm is required to detect anomalous frames in testing. In the past few years, researchers have put lots of effort into enhancing video anomaly detection performance on standard datasets: UCSD Pedestrian~\cite{Mahadevan2010ped}, CUHK Avenue~\cite{Lu2013Avenue} and ShanghaiTech Campus dataset~\cite{liu2018future}. A popular branch of approach is the reconstruction-based methods which employ autoencoders for normal training data reconstruction and utilize reconstruction error as the anomaly score for test-time anomalous frame discrimination. For example,~\cite{hasan2016learning} takes advantage of a deep convolutional network for spatio-temporal reconstruction-based VAD.~\cite{luo2017remembering} incorporates convolutional Long Short-Term Memory~\cite{hochreiter1997long} network into autoencoder for better consecutive frame temporal relationship modeling. In addition,~\cite{gong2019memorizing} integrates memory with autoencoders for enhanced performance. Another prevailing type of method is the prediction-based approach. The rationale is that the current frame is generated using preceding frames and the predictions with large error w.r.t. the true frames are considered anomalous. This concept is first proposed in~\cite{liu2018future} where U-Net~\cite{ronneberger2015u} is adopted as the architectural framework for frame prediction. A natural idea is to mix reconstruction-based techniques with prediction-based ones, resulting in hybrid methods~\cite{tang2020integrating,Liu_2021_HFVAD}. More recently, the approaches based on self-supervised learning have begun to dominate in VAD performance~\cite{Georgescu2021ssmtl,ristea2022self}. Specifically,~\cite{Georgescu2021ssmtl} manually defines four pretext tasks for efficient learning of normal data patterns and~\cite{ristea2022self} proposes a masked dilated filter equipped with self-supervised reconstruction loss.

\begin{table*}
    \centering
    \caption{Summary of explainable methods in 2D anomaly detection. ``Type" represents in which way the method provides explanations. ``Det." and ``Loc." indicate whether the method can do anomaly localization (providing pixel-level scores) or simply can do detection (providing sample-level scores). For ODD works, we only list the testing out-of-distribution datasets.}
    \renewcommand{\arraystretch}{1.2}
    \setlength{\tabcolsep}{2.5pt}
    \begin{tabular}{c|c|c|l|cc|l}
    \toprule
        \textbf{Modality} & \textbf{Type} & \textbf{Year} & \multicolumn{1}{c}{\textbf{Methods}}  & \textbf{Det.}& \textbf{Loc.} & \multicolumn{1}{c}{\textbf{Datasets}} \\ \midrule
        \multirow{22}{*}{Image} & \multirow{2}{*}{Attention}& 2020&  GradCon~\cite{kwon2020backpropagated}  & \checkmark&  &C10, MN, fMN \\ 
        & & 2021&  FCCD~\cite{liznerski2021explainable}   & \checkmark &\checkmark & fMN, C10, ImageNet, CURE-TSR \\ \cdashline{2-7}
        & \multirow{4}{*}{Input perturbation}& 2018&  ODIN~\cite{liang2018enhancing} & \checkmark& & TinyImageNet, LSUN, iSUN  \\ 
        & & 2018&  Mahalanobis~\cite{lee2018simple}  & \checkmark& & C10, SVHN, ImageNet, LSUN \\ 
        & & 2020&  Generalized ODIN~\cite{Hsu_2020_CVPR}  & \checkmark& & TinyImageNet, LSUN, iSUN, SVHN  \\ 
        & & 2022&  SLA$^2$P~\cite{wang2022self} & \checkmark& & C10, C100, Cal \\ \cdashline{2-7}
        & \multirow{9}{*}{Generative model} & 2017&  AnoGAN~\cite{schlegl2017unsupervised} & \checkmark& & Retina OCT \\ 
        & & 2018&  ALAD~\cite{zenati2018adversarially} & \checkmark& & MN \\ 
        & & 2019&  f-AnoGAN~\cite{schlegl2019f} & \checkmark& \checkmark & UKB, MSLUB, BRATS, WMH  \\ 
        & & 2019&  Genomics-OOD~\cite{ren2019likelihood} & \checkmark& & MN, SVHN \\ 
        & & 2020&  Likelihood Regret~\cite{xiao2020likelihood} & \checkmark& & MN, fMN, C10, SVHN, LSUN, CelebA\\ 
        & & 2021&  FastFlow~\cite{yu2021fastflow} & \checkmark& \checkmark& MVTAD, BTAD, C10\\
        & & 2022&  AnoDDPM~\cite{wyatt2022anoddpm} & \checkmark& \checkmark & MVTAD \\ 
        & & 2022&  Diffusion-anomaly~\cite{wolleb2022diffusion} & \checkmark& \checkmark & BRATS\\ 
        & & 2022&  DDPM~\cite{pinaya2022fast}&  \checkmark& \checkmark & MedMNIST, UKB, MSLUB, BRATS, WMH  \\ 
        & & 2023&  DiAD~\cite{he2023diad}&  \checkmark& \checkmark & MVTAD, VisA, BRATS  \\ 
        & & 2025&  DeCo-Diff~\cite{anonymous2025correcting} & \checkmark& \checkmark & MVTAD, VisA\\
        \cdashline{2-7}
        & \multirow{7}{*}{Foundation model} & 2023&  WinCLIP~\cite{jeong2023winclip} & \checkmark& \checkmark & MVTAD, VisA \\ 
        & & 2023&  CLIP-AD~\cite{chen2023clip} & \checkmark& \checkmark & MVTAD, VisA \\
        & & 2023&  AnomalyCLIP~\cite{zhou2023anomalyclip} & \checkmark& \checkmark & MVTAD, VisA, MPDD, BTAD, SDD, DAGM \\
        & & 2023&  SAA+~\cite{cao_segment_2023} &  \checkmark& \checkmark & MVTAD, VisA, KSDD2, MTD\\
        & & 2023&  AnomalyGPT~\cite{gu2023anomalyagpt} &  \checkmark& \checkmark & MVTAD, VisA\\ 
        & & 2023&  Myriad~\cite{li2023myriad} &  \checkmark& \checkmark & MVTAD, VisA\\ 
        & & 2023&  GPT-4V~\cite{cao2023towards} &  \checkmark& & MVTAD, VisA, MVTLC, Retina OCT \\
        & & 2023&  GPT-4V-AD~\cite{zhang2023exploring} &  \checkmark& \checkmark & MVTAD, VisA\\
        & & 2024&  CLIP-SAM~\cite{li2024clipsam} &  \checkmark& \checkmark & MVTAD, VisA\\
        & & 2024&  AnomalyClip~\cite{anonymous2024anomalyclip} &  \checkmark& \checkmark & MVTAD, VisA\\
        & & 2024&  AA-CLIP~\cite{anonymous2025aaclip} &  \checkmark& \checkmark & MVTAD, VisA, MPDD, BTAD, Retina OCT\\
        & & 2024&  MVFA~\cite{huang2024adapting} &  \checkmark& \checkmark & BrainMRI \\
        & & 2024&  InCTRL~\cite{zhu2024generalist} &  \checkmark& \checkmark & MVTAD, VisA, BTAD MPDD, BrainMRI \\
        & & 2025&  Bayes-PFL~\cite{anonymous2025bayesian} &  \checkmark& \checkmark & MVTAD, VisA, KSDD2, BTAD, DAGM, BrainMRI \\
        & & 2025&  LogSAD~\cite{zhang2025trainingfree} &  \checkmark& \checkmark & MVTAD, VisA, MVTLC \\
        & & 2025&  LogicAD~\cite{jin2025logicad} &  \checkmark&  & MVTLC \\
        & & 2025&  Anomaly-OV~\cite{anonymous2025zeroshotreasoning} &  \checkmark& \checkmark & MVTAD, VisA, BTAD, MPDD, BrainMRI \\
        & & 2025&  UniVAD~\cite{anonymous2025zeroshotreasoning} &  \checkmark& \checkmark & MVTAD, VisA, MVTLC, BrainMRI \\
        \midrule
        \multirow{4}{*}{Image+Video} & \multirow{3}{*}{Attention} & 2020&  AD-FactorVAE~\cite{liu2020towards}   & \checkmark& \checkmark& MN, MVTAD, Ped \\ 
        & & 2020&  CAVGA~\cite{venkataramanan2020attention}  & \checkmark& \checkmark & MVTAD, mSTC, LAG, MN, C10, fMN  \\ 
        & & 2022&  SSPCAB~\cite{ristea2022self}& \checkmark& \checkmark& MVTAD, Ave, STC  \\ \cdashline{2-7}
        & \multirow{2}{*}{Generative model} & 2019&  LSAND~\cite{abati2019latent} & \checkmark& \checkmark& MN, C10, Ped, STC\\
        & & 2022&  CFLOW-AD~\cite{gudovskiy2022cflow} & \checkmark& \checkmark& MVTAD, STC\\
        \midrule
        \multirow{11}{*}{Video} & \multirow{2}{*}{Attention} & 2020&  Self-trained DOR~\cite{pang2020self} & \checkmark & \checkmark & Ped2, Subway, UMN \\
        & & 2021&  DSA~\cite{purwanto2021dance} & \checkmark & \checkmark & UCF, STC\\ \cdashline{2-7}
        & \multirow{9}{*}{Reasoning} & 2018&  Scene Graph~\cite{chen2018scene}&\checkmark&  & UCF \\ 
        & & 2021&  CTR~\cite{wu2021learning} & \checkmark  &  & UCF, STC, XD \\ 
        & & 2023&  Interpretable~\cite{doshi2023towards} &\checkmark & \checkmark  & Ave, STC\\ 
        & & 2024&  VADor w LSTC~\cite{lv2024video} &\checkmark & & UCF, TAD \\ 
        & & 2024&  VERA~\cite{ye2024vera} &\checkmark & & UCF, XD \\
        & & 2024&  AnomalyRuler~\cite{yang2024follow} &\checkmark & & Ped2, Ave, STC, UBnormal  \\
        & & 2024&  LAVAD~\cite{zanella2024harnessing} &\checkmark & & UCF, XD  \\
        & & 2024& Holmes-VAU~\cite{zhang2024holmes} &\checkmark & \checkmark & UCF, XD  \\
        & & 2025& Anomize~\cite{li2025anomize} &\checkmark & & UCF, XD  \\
        \cdashline{2-7}
        & \multirow{5}{*}{Intrinsic interpretable} & 2017&  JDR~\cite{hinami2017joint}&\checkmark  & \checkmark  & Ped2, Ave\\
        & & 2021&  XMAN~\cite{szymanowicz2021x} &\checkmark & \checkmark  & X-MAN\\ 
        & & 2022&  VQU-Net~\cite{szymanowicz2022discrete} &\checkmark & \checkmark  & Ped2, Ave, X-MAN \\ 
        & & 2022&  AI-VAD~\cite{reiss2022attribute} &\checkmark &   & Ped2, Ave, STC\\ 
        & & 2022&  EVAL~\cite{singh2022eval} &\checkmark  & \checkmark  & Ped1, Ped2, Ave, STS, STC\\ \cdashline{2-7} 
        & \multirow{2}{*}{Memory} & 2020&  Memory AD~\cite{park2020learning}&\checkmark  & \checkmark  & Ped2, Ave, STC\\
        & & 2021&  DLAN-AC~\cite{yang2022dynamic} &\checkmark &  & Ped2, Ave, STC\\ 
        \bottomrule
    \end{tabular}
    \label{tab: main}
\end{table*}
\section{Explainable 2D Visual Anomaly Detection}
\label{sec: 3-methods}
Explainable AI is essential for improving trust and transparency in AI-based systems~\cite{adadi2018peeking}. In anomaly detection, explainability is an ethical and regulatory requirement, particularly in safety-critical domains~\cite{10.1145/3609333,ramachandra2020survey}. We provide a detailed overview of methods that incorporate explainability into 2D visual anomaly detection, categorized by their primary approach and modality.
\subsection{Explainable Image Anomaly Detection} \label{sec: x-iad}
\begin{figure}
    \centering
    \includegraphics[width=0.9\linewidth]{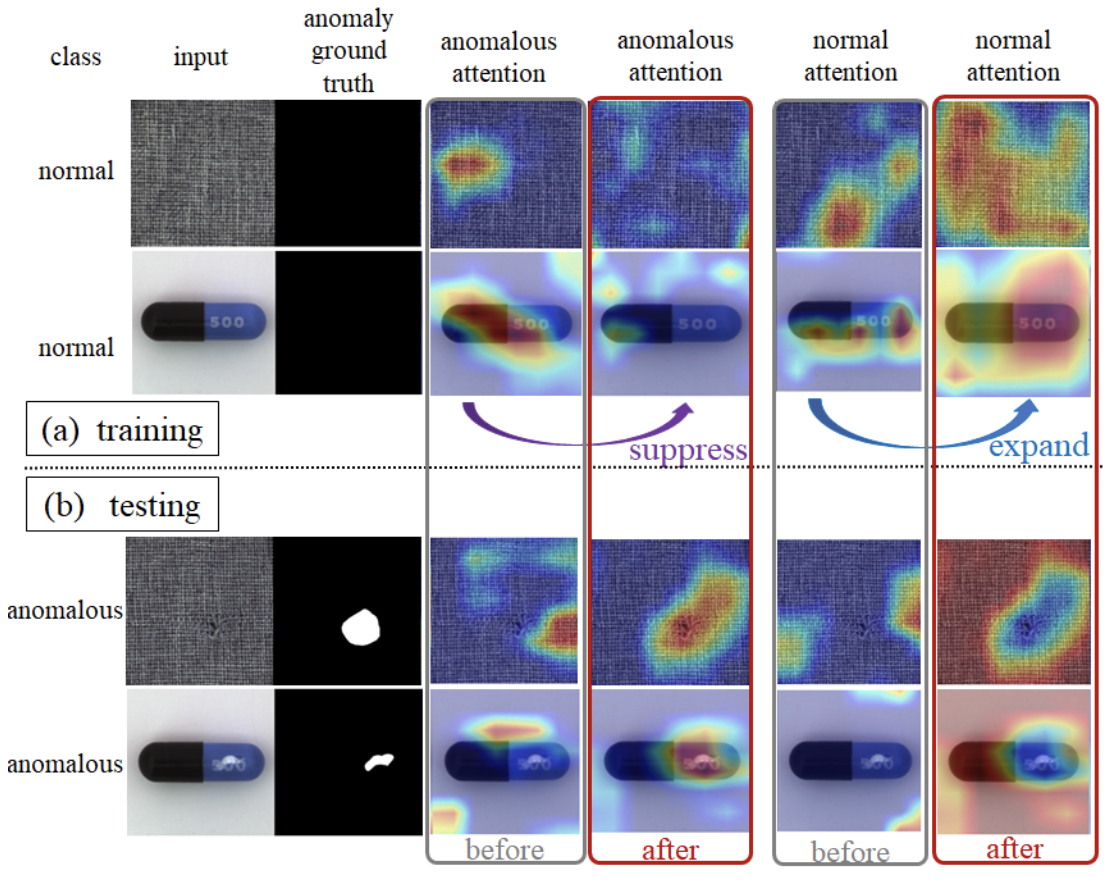}
    \caption{\cite{venkataramanan2020attention}. The proposed CAVGA approach employs the complementary guided attention loss to force the attention map to cover all the normal regions and meanwhile suppress the anomalous attention on the normal images.}
    \label{fig: attention-iad}
    \vspace{-2mm}
\end{figure}
For image anomaly detection, either semantic anomaly detection or pixel-level anomaly segmentation, explainability, and interpretability are of great significance to the understanding of why specific samples are considered anomalies and the development of more advanced approaches. Explainable image anomaly detection techniques can be broadly grouped into three kinds: Attention-based IAD, Input-perturbation-based IAD, and Generative-model-based IAD. 



\subsubsection{\bf Attention-based IAD}
Attention in image recognition mainly refers to the concept that the model regards some parts or locations of a map or image as more important or interesting. Attention can be 1) post hoc (calculated with a trained model in hand), or 2) learned in a data-driven way. For the first type, the generated attention maps can be directly used to explain anomalies, as the high activation areas correspond to locations that the model pay more attention to and therefore indicate more abnormality. For instance,~\cite{liu2020towards} generates attention from the latent space of a variational auto-encoder (VAE) through gradient calculation. Such attention maps can be adopted to detect anomalies in images. Also,~\cite{liznerski2021explainable} develops a Fully Convolutional Data Description method that maps the samples to an explanation attention map. There are some recent works that further take advantage of the post hoc attention to improve training.~\cite{venkataramanan2020attention} makes use of the gradient-based attention to regularize the normal data training of the proposed convolutional adversarial VAE by proposing an attention expansion loss to enforce the attention map to concentrate on all normal parts. When having few anomalous training images in hand, they further put forward a complementary guided attention loss regarding the proposed binary classifier to refrain the network from focusing on erroneous areas. As illustrated in Fig.~\ref{fig: attention-iad}, such practice makes the network spare no effort to encode the normal patterns of the training images and feedback-intensive responses in anomalous areas of images in the testing phase, accounting for why it outperforms the plain variants of autoencoders. Such explainable image anomaly detection methods usually compute the attention scores using functions similar to the Grad-CAM~\cite{8237336}, taking the form of 
\begin{equation}
    a_k = \frac{1}{M}\sum_i\sum_j \frac{\partial y}{\partial A_{ij}^k},
\end{equation}
where $y$ can be any differentiable activation, $A^k$ denotes the activations of feature map $k$, $M$ is the total number of pixels in the feature map, i.e.
\begin{equation}
    M = \sum_i \sum_j 1.
\end{equation}
~\cite{kwon2020backpropagated} takes the first attempt to exploit backpropagated gradients as representations to design anomaly score function and puts forward a novel directional gradient alignment constraint as part of the training loss function. For the second type of attention,~\cite{ristea2022self} introduces a channel attention mechanism after a masked convolutional layer for the reconstruction neural network design of the masked part of the convolutional receptive field. The proposed channel attention module is able to aggregate and recalibrate the feature information and selectively suppress reconstruction maps.

\subsubsection{\bf Input-perturbation-based IAD}

\begin{table*}[htbp]
\centering
\caption{Summary of the detailed perturbation formula in input-perturbation-based image anomaly detection task. $\epsilon$ denotes the perturbation magnitude, $\text{sign}$ denotes the signum function, $S(x)$ is some score function of $x$ and $D_{\text{in}}^{\text{val}}$ denotes the in-distribution validation set. The dashed line means that the first three methods utilize pre-trained classifier output for $S(x)$ while the last one uses a self-supervised learned classifier.}
\scalebox{0.95}{
\begin{tabular}{lccr}
\toprule
 {\bf Method}  & {\bf Perturbation term} & {\bf Magnitude $\epsilon$} & {\bf Score $S(x)$} \\ \hline
 ODIN~\cite{liang2018enhancing}& $-\epsilon \text{sign} (-\nabla_x \log S(x))$ & searched& maximum softmax probability  \\
Mahalanobis~\cite{lee2018simple}& $+\epsilon \text{sign}(\nabla_{x} S(x))$ & searched& Mahalanobis distance-based confidence score\\
 Generalized ODIN~\cite{Hsu_2020_CVPR}& $-\epsilon \text{sign} (-\nabla_x S(x))$ & $\argmax_{\epsilon}\sum_{x \in D_{\text{in}}^{\text{val}}} S(x)$& maximum softmax probability\\ \hdashline
 SLA$^2$P~\cite{wang2022self}& $-\epsilon \nabla_{x} \log S(x)$ & searched& maximum softmax probability \\
\bottomrule
\end{tabular}
}
\label{tab: perturbation}
\end{table*}

In image-level anomaly detection tasks, input perturbation is a commonly adopted technique in both semantic anomaly detection (focusing on category difference)~\cite{wang2022self} and out-of-distribution detection (focusing on dataset-wise distribution shift)~\cite{liang2018enhancing,Hsu_2020_CVPR,lee2018simple}. In semantic anomaly detection, the training phase includes only data from a single class, termed the positive class. However, during the inference phase, the classifier deals with data from the positive class as well as data outside of it, which is sometimes called the negative class. The primary goal of semantic anomaly detection is to ascertain whether an item under examination belongs to the class that was represented during the training process~\cite{perera2021one,khan2010survey}. So basically the difference between anomalies and normal data lies in the semantic shift (i.e., they are from different categories). On the contrary, OOD detection aims to detect anomalies with covariate shift~\cite{yang2021oodsurvey}, which means that the anomalous data are from different domains from the inliers. So in OOD detection, the standard benchmark setting is the training data is from one dataset (e.g. CIFAR-10) and the testing data is mixed by the training dataset and a new dataset from a different domain (e.g., CIFAR-10 and CIFAR-100). The OOD detection algorithm is expected to detect the data instances from the new dataset out. For both OCC and OOD detection, the input perturbation can be generally formalized as 
\begin{equation}
    x' = x + \text{perturbation term},
\end{equation}
where $x$ denotes the input data or feature, and $x'$ is the perturbed data or feature. The perturbation term is usually in the form of a perturbation magnitude multiplying the gradient of some score function w.r.t. the input. Then the perturbed data or feature serves as input for anomaly score calculation.~\cite{liang2018enhancing} is the pioneering work adopting input perturbation technique in anomaly detection. The proposed ODIN method takes the gradient of a pretrained network and aims to increase the softmax score of any given input.~\cite{Hsu_2020_CVPR} generalizes ODIN via finding the maximum magnitude which maximizes the sum of the scores w.r.t. the in-distribution validation dataset. The generalized ODIN approach further enhances detection performance consistently. Instead of directly using softmax output scores,~\cite{lee2018simple} proposes Mahalanobis distance-based confidence score, and the perturbation is introduced to enhance the confidence score. Different from these OOD detection work which computes the gradient using a pretrained network,~\cite{wang2022self} trains a neural network with feature-level self-supervised learning and adds perturbation on the feature embeddings using the trained classifier network. The detailed comparison of the input perturbations of these literature is summarized in Tab.~\ref{tab: perturbation}. Despite the differences among these input-perturbation methods, the rationale that they can explain anomalous patterns is the same: due to the characteristics of the anomaly detection problem anomalies are lower in quantity compared to the normal samples and the modes of anomalies are more diverse when applying the same amount of perturbation, the score alternations of them will be different from those of the normal data. Therefore we are able to examine the effect of output via input changes to generate reasonable explanations.

\subsubsection{\bf Generative-model-based IAD}

Generative models (GM) are widely adopted in 2D anomaly detection. The most commonly used GMs include Generative Adversarial Networks (GAN), Autoregressive Models, Variational Autoencoders (VAE),  Normalizing Flows, Diffusion Models and other variants of autoencoders.~\cite{schlegl2017unsupervised} is the first work to seamless combine GAN with image anomaly detection and localization task, where the reconstruction error of GAN and the discrimination score from the discriminator are jointly used for anomaly score design. This is because the GAN fails to reconstruct the anomalies well and the discriminator will regard the anomalies as fake since they only encounter normal ones in the training phase.~\cite{abati2019latent} employs a deep autoencoder together with an autoregressive model for density estimation which learns the distribution of the latent representations via maximum likelihood principles and proves empirically this is an efficient regularizer for anomaly detection. Furthermore,~\cite{ren2019likelihood} demonstrates that likelihood from generative model-based anomaly detection methods is problematic due to the background statistics and proposes a novel likelihood ratio-based anomaly score. The key finding is that likelihood value is mainly controlled by the background pixels while likelihood ratios concentrate more on the semantic information, which explains why likelihood ratio is a more preferable choice for anomaly score for one-class classification. The formulation of the proposed likelihood ratio is
\begin{equation}
    \text{LLR}(x) = \log p_\theta (x) - \log p_{\theta_0} (x),
\end{equation}
where $x$ is the data sample, $p_\theta(\cdot)$ is the model trained employing in-distribution data and $p_{\theta_0}(\cdot)$ is a background model which captures general background statistics. Similarly,~\cite{xiao2020likelihood} addresses the problem that generative models, especially VAEs, are inclined to assign higher likelihoods on some types of OOD samples through defining a new OOD detection score dubbed as Likelihood Regret. They show that the Likelihood regret can be viewed as a specific form of log ratio of the likelihood from VAE. It is effective in separating anomalies out in that the increase of likelihood by substituting the model configuration with the optimal one for the single sample ought to be comparatively small for normal samples and large for anomalous ones thanks to the fact that the Variational AE is trained well on the normal training data.
\begin{equation}
    \text{LR}(x) = L(x; \theta^*, \hat{\tau}(x)) - L(x; \theta^*, \phi^*).
\end{equation}
Here 
\begin{equation}
    L(x;\theta,\phi) = \EE_{q_\phi(z|x)}[\log p_{\theta} (x|z)] - D_{\text{KL}}[q_{\phi}(z|x) \| p(z)]
\end{equation}
represents the evidence lower bound, (ELBO), $\phi$ and $\theta$ are the parameters of the encoder and decoder respectively, $\phi^*$ and $\theta^*$ are the network parameters maximizing the population log-likelihood over the training dataset, and $\hat{\tau}(x)$ denotes the optimal parameter maximizing the ELBO of the test input $x$.
~\cite{liu2020towards} also utilizes VAE for anomaly detection, but instead uses a gradient-based attention map generated from the latent space of VAE for anomaly detection and localization in images. FastFlow~\cite{yu2021fastflow} employs 2D normalizing flows for efficient and unsupervised anomaly detection and localization, providing explainability through intuitive visualizations of the transformed features for anomaly identification. CFLOW-AD~\cite{gudovskiy2022cflow} leverages conditional normalizing flows for unsupervised anomaly detection, offering explainability through the generative model's ability to localize anomalies by contrasting reconstructed normal and anomalous regions, thereby enhancing the interpretability of detection results. More recently, diffusion models exhibit superior image generation ability and hence emerge in anomaly detection research. For example, AnoDDPM~\cite{wyatt2022anoddpm} proposes a partial diffusion strategy with simplex noise for medical image anomaly detection and localization.~\cite{wolleb2022diffusion} integrates DDPM~\cite{ho2020denoising} with a binary classifier for medical image anomaly segmentation and~\cite{pinaya2022fast} devises a diffusion model for fast brain image anomaly segmentation. More recently, DiAD~\cite{he2023diad} develops a framework for detecting anomalies across multiple classes, integrating an autoencoder in pixel space, a semantic-guided network in latent space connected to the denoising network of the diffusion model, and a pre-trained extractor in the feature space, with innovations like spatial-aware feature fusion for enhanced reconstruction accuracy and semantic preservation. Further exploration involves reformulating the diffusion process itself to better correct deviations from normality for multi-class unsupervised anomaly detection~\cite{anonymous2025correcting}. The critical advantage of the Diffusion Model for explainability is that each step in the forward step and reverse step has a closed form, which enables people to visualize the latent outputs of each step or layer in order to investigate how anomalies are ruled out in the diffusion process and where the model pay more attention.

\begin{figure}[tbp]
    \centering
    \includegraphics[width=0.9\linewidth]{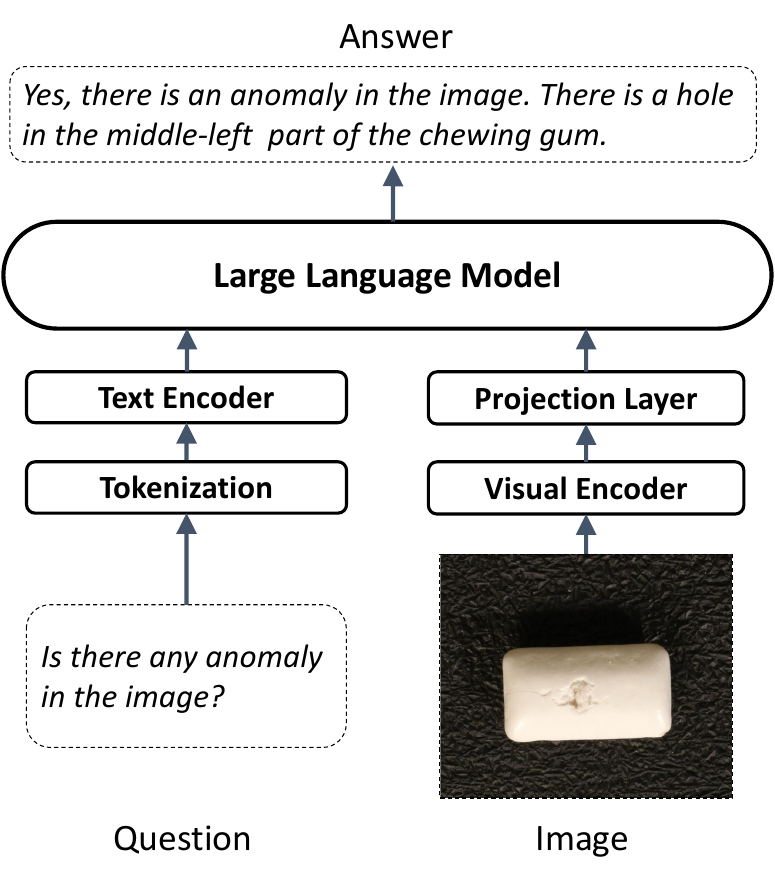}
    \caption{Illustration of LLM-based image anomaly detection. A projection layer is usually applied after the extracted visual features to project them into the input token space of the LLM.}
    \label{fig: llm-ad}
\end{figure}

\subsubsection{\bf Foundation-model-based IAD}
With the recent rise of foundation model architectures, there exist several very recent works in image anomaly detection literature leveraging the extraordinary reasoning ability of large foundation models trained with massive amounts of data. A series of work has explored adapting CLIP~\cite{radford2021learning} for few-shot or zero-shot image anomaly detection and localization, leveraging its aligned vision-language representations. Examples include CLIPAD~\cite{chen2023clip}, AnomalyCLIP~\cite{zhou2023anomalyclip}, WinCLIP~\cite{jeong2023winclip}, and CLIP-SAM~\cite{li2024clipsam}. Recent advancements focus on enhancing zero-shot capabilities through object-agnostic prompt learning~\cite{anonymous2024anomalyclip}, anomaly-aware adaptations of CLIP~\cite{anonymous2025aaclip}, and Bayesian approaches to prompt learning~\cite{anonymous2025bayesian}. The basic idea is to utilize the alignment property and extraordinary representation ability of text and visual features of CLIP for generic abnormality learning and discrimination, which makes the methods naturally explainable in the corresponding text prompt space. Some approaches incorporate learnable class names to aid detection, particularly in long-tailed scenarios~\cite{ho2024longtailed}. Further, Segment Any Anomaly~\cite{cao_segment_2023} introduces the "Segment Any Anomaly + (SAA+)" framework, a novel approach for anomaly segmentation under the zero-shot setting that utilizes hybrid prompt regularization to improve the adaptability of foundational models. Instead of relying on domain-specific fine-tuning like traditional anomaly segmentation models, SAA+ leverages the zero-shot generalization capabilities of foundational models, assembling them to utilize diverse multi-modal prior knowledge for anomaly localization, and further adapts them using hybrid prompts derived from domain expert knowledge and target image context. The method's explainability stems from its innovative use of hybrid prompts, which are derived from both domain expert knowledge and the specific context of the target image. Specific adaptations have also been developed for generalizable anomaly detection in medical images using VLMs~\cite{huang2024adapting}.

Nowadays, with the rapid development of LLMs and Large MultiModal Models (LMMs)~\cite{yang2023dawn}, the outstanding reasoning ability of pretrained language models has been employed for enhanced anomaly detection and explanations. AnomalyGPT~\cite{gu2023anomalyagpt} leverages LVLM for Industrial Anomaly Detection by simulating anomalous images with textual descriptions and fine-tuning LVLMs with prompt embeddings for direct anomaly presence and location assessment without manual thresholds. This approach is explainable as it supports multi-turn dialogues, allowing users to engage with the model for detailed anomaly localization and understanding, facilitated by a lightweight decoder for fine-grained semantic capture and prompt learner integration for enhanced LVLM alignment. Myriad~\cite{li2023myriad} introduces a novel large multimodal model, leveraging vision experts for enhanced industrial anomaly detection (IAD), which includes an expert perception module for embedding prior knowledge from vision experts and an expert-driven vision-language extraction module for domain-specific vision-language representation. This method enhances explainability by incorporating expert-generated anomaly maps into the learning process, thereby providing detailed descriptions of anomalies, which improves both detection accuracy and the comprehensibility of the detection process;~\cite{cao2023towards} tests the GPT-4V~\cite{yang2023dawn} capabilities in generic anomaly detection via examples, and~\cite{zhang2023exploring} adapts GPT-4V for zero-shot anomaly detection in images through granular region division, prompt designing, and text to segmentation, enhancing visual question answering for direct anomaly evaluation without prior examples. Its explainability stems from leveraging VQA to interpret and segment anomalies based on textual prompts and image analysis, providing intuitive and detailed anomaly identification and localization capabilities. More recent works continue this trend, exploring training-free approaches~\cite{zhang2025trainingfree}, employing VLM-based text features for logical explanation~\cite{anonymous2025logicad}, and leveraging MLLMs for explicit zero-shot reasoning about anomalies~\cite{anonymous2025zeroshotreasoning}. All these existing LLM/LMM-based IAD methods are typically structured by concatenating extracted text and image embeddings and feeding them into an LLM, as depicted in Fig.~\ref{fig: llm-ad}. Given a question related to the anomaly, the LLM can determine whether the image contains any anomaly, where it is, detailed descriptions, and even explanations. Owing to the massive knowledge the LLM is trained on, it can naturally generate meaningful textual explanations, and further clarifications can be acquired by changing or adding prompts. The integration of few-shot learning techniques with residual learning also shows promise for building generalist anomaly detection systems with potential explainability benefits~\cite{zhu2024generalist}, as does the development of unified models for few-shot detection~\cite{anonymous2025univad}.
\subsection{Explainable Video Anomaly Detection} \label{sec: x-vad}
Explainability is a critical component in video anomaly detection. Currently, there is no particular definition of explainability in video anomaly detection. Also, it is not quantifiable. Thus, existing methods explain themselves by providing some high-level ideas, such as anomaly localization, and action recognition, and some of them are intrinsically interpretable. In this section, we divide them into three main categories: Atention-based methods, Reasoning-based methods, and Intrinsic interpretable methods. 

\subsubsection{\bf Attention-based methods}  
\begin{figure}[tbp]
    \centering
    \includegraphics[width=\linewidth]{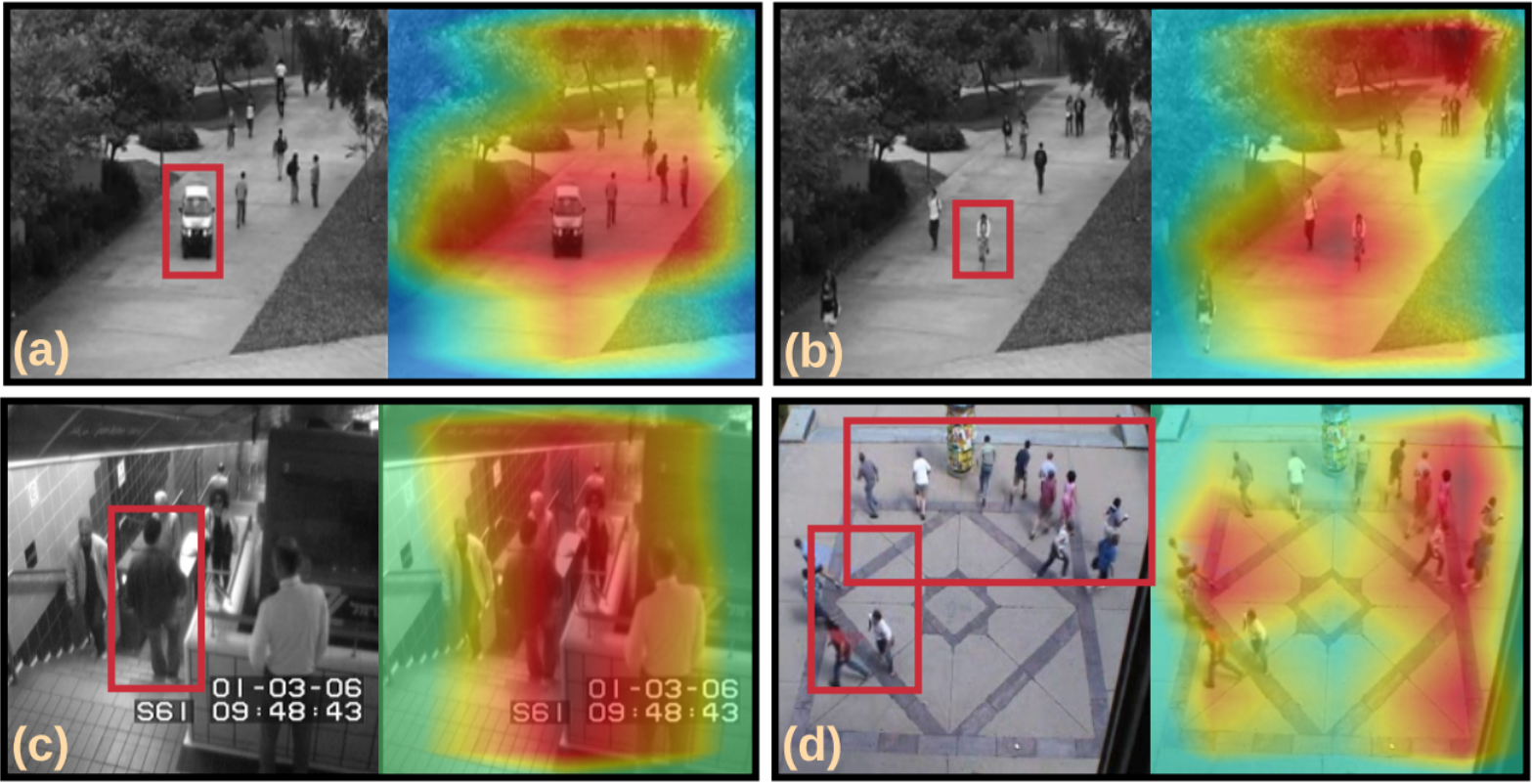}
    \caption{Image from \protect\cite{pang2020self}. Manually labeled anomalies in the original inputs (red rectangles on the left) and the corresponding CAM based saliency maps yielded by their method (right).}
    \label{fig:attention}
\end{figure}
Attention-based methods are low-level explainable methods that give human insights into what a model is looking at. In other words, they highlight the pixels that contribute the most to the prediction. According to the attention map they produce, we can have a better idea of what specific areas do the model mainly focuses on and whether the area is aligned with human prior knowledge. In this case, it is very easy to troubleshoot the errors that the model makes.  
The typical process of attention-based methods is first to calculate the scores for anomalies in each frame, and then the attention-based method traces back from the anomaly score prediction to the activation map of raw frames. Specifically, the anomaly score estimation can be concluded as a sequential task, including the frame feature encoder and anomaly score function. Here we can formulate them as $ S = f_\theta(X)$, where $S$ indicates the anomaly scores and $X = \{x_1, x_2, \cdots\}$ represents the sequential frames, and $f_\theta(\cdot)$ means the combination of feature encoder and the anomaly score function. 
Calculating the anomaly scores can be done in various ways. However, to achieve attention-based interpretability, the key is to adopt the activation map from the anomaly scores. In the context of the given frame \(x\), the \(p_{k}(i, j)\), denoting the activation value of a unit indexed by \(k\) within the last layer, localized at the spatial coordinates \((i, j)\). Concurrently, the symbol \(w_{k}\) is designated to represent the weight linked to unit \(k\) in relation to the computation of the anomaly score. $\phi(x)$ is defined as the attention score of a frame which can be expressed as:
\begin{equation}
    \phi(x) = \sum_{i,j} \sum_{k} w_{k} \cdot p_{k}(i, j).
\end{equation}
This formulation encapsulates the interplay between the activation values, their associated weights, and the resultant anomaly score, forming a fundamental construct within the framework.

The key idea of attention-based methods is to capture the gradient. How to calculate the gradient and how to use them to explain the model are the novel parts of this kind of method.~\cite{pang2020self} used a simple approach, class activation mapping (CAM)~\cite{zhou2016learning}, to calculate attention weights. They then localized the identified anomaly by looking at high activation values on the attention map.~\cite{purwanto2021dance} used a self-attention mechanism to model the relationships of the nodes in spatio-temporal graphs. They construct conditional random fields as nodes and learn self-attention on them. Visualization of the heat map on the self-attention random fields reveals the location of the anomaly. Other papers, such as~\cite{venkataramanan2020attention}, also use attention to localize the anomaly and give an explanation based on that.
Although the attention map indicates the importance of a specific region or features that play a significant role in the prediction, it cannot explain why a model makes a correct or wrong prediction. For example, in Fig.~\ref{fig:attention}, the attention map indicates a larger region than the anomaly, which reveals little information. Also, as~\cite{cao2022random} discussed, a random ResNet can provide an attention map that can capture the foreground object. In this case, it can tell nothing even if it gives an accurate prediction.

\subsubsection{\bf Reasoning-based methods}
\begin{figure}
    \centering
    \includegraphics[width=\linewidth]{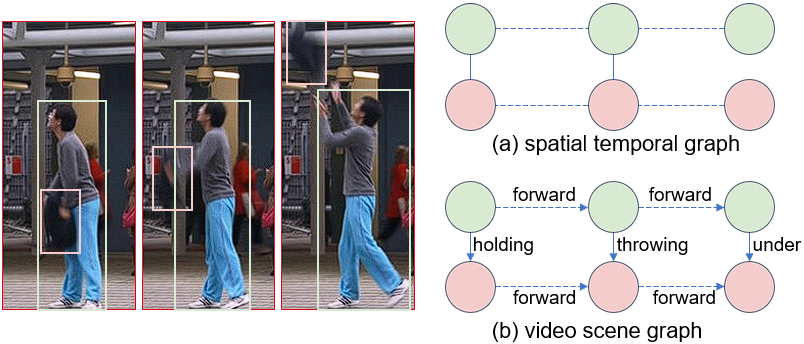}
    \caption{Illustration of (a) spatial-temporal graph and (b) video scene graph generation using consecutive frames from the Avenue dataset. Both graphs capture valuable semantic information regarding objects. In the spatial-temporal graph, objects with overlapping bounding boxes are connected by spatial edges. Temporal edges indicate consecutive frames for a given object, effectively incorporating temporal information. These edges also encode semantic details, such as bounding box distances. In the case of video scene graphs, (subject, predicate, object) triplets are employed to articulate relationships between objects. This enables the description of frames in a semantically meaningful manner. For instance, (man, hold, bag) represents the initial frame, which then transitions to (man, throw, bag), and eventually (man, under, bag). These changing relations offer a high-level explanation of anomalies within the scene.}
    \label{fig:scene_graph}
\end{figure}

\paragraph{Logic-based methods}
Different from the attention-based method, logic-based methods focus on giving high-level explanations by providing some context reasoning techniques, such as knowledge graph, scene graph, and causal inference. They explain the model in a semantic way that considers the foreground objects and the background information together. As shown in Fig.~\ref{fig:scene_graph}, the frame information can be formulated as graphs, including spatial-temporal graph, knowledge graph, scene graph, and so on. The information provided by graph can not only benefit anomaly detection but also bring interpretability to models.

For example,~\cite{doshi2023towards} proposed a framework with a global monitor and a local monitor. The global monitor constructs a scene graph that captures the global information, and the local monitor focuses on the human pose. Each interaction is observed by calculating its semantic distance from a collection of nominal embeddings derived from the training data. Typically, the (subject, predicate, object) triplet is mapped to numerical vectors using a semantic embedding network. The goal of metric learning is to extract information regarding the absence or presence of anomalies from the interaction embeddings. During training, the positive instance is randomly selected from the nominal training set, and the negative instance is randomly sampled from an artificially generated set of anomalous interaction triplets, e.g., a person hits a person. The objective is to minimize the distance between the positive samples and to maximize the negative samples. Thus, during testing, when an anomaly occurs, it can give the true reason for the anomaly and output an interpretable result. Similarly,~\cite{chen2018scene} also relied on scene graphs to interpret the anomaly. However, it took advantage of multiclass SVM and naive Bayesian for anomaly detection. 

The knowledge graph can also capture semantic information.~\cite{nesen2020knowledge} constructed a knowledge graph by aligning prior knowledge and recognized objects, which provides a semantic understanding of a video. For example, a supercar in a poor district is an anomaly event. Techniquely, the relationship between the ontologies of the knowledge graph and objects detected in a video is determined by calculating the similarity between entities present in the frame. In addition, causal inference is born to explain models.~\cite{wu2021learning} proposed a module for causal temporal relations to capture the temporal cues between video snippets, which simultaneously explores two types of temporal relationships: semantic similarity and positional prior.
Reasoning-based methods provide high-level explanations that are similar to human understanding. However, it is more complicated and requires high-quality object recognition model in the first place. In other words, it will fail if an object is misclassified.

\paragraph{LLM-based methods}
LLM-based methods improve explainable video anomaly detection via integrating visual and linguistic information. VERA~\cite{ye2024vera} employs learned reflective questions to guide vision-language models in generating anomaly scores, ensuring explainability through these questions. AnomalyRuler~\cite{yang2024follow} detects anomalies through inducing rules from normal samples, with the rules themselves providing clear explanations. LAVAD~\cite{zanella2024harnessing} generates textual descriptions and leverages LLMs for reasoning, offering explanations via the descriptions and prompting process. Holmes-VAU~\cite{zhang2024holmes} targets anomaly-rich regions using a multimodal LLM, delivering explanations at varying levels of granularity. VadCLIP aligns visual and linguistic features for precise detection, linking anomalies to descriptive language for explainability. Anomize~\cite{li2025anomize} guides detection with textual information and label relations, explaining novel anomalies through these connections.~\cite{lv2024video} produces textual explanations using video-based LLMs, making the detection process transparent. These methods collectively offer diverse, explainable approaches to identifying anomalies in video data harnessing the strong reasoning ability of the LLM backbones which were pre-trained on a large amount of textual data.

\subsubsection{\bf Intrinsically interpretable methods}
\begin{figure}[tbp]
    \centering
    \includegraphics[width=\linewidth]{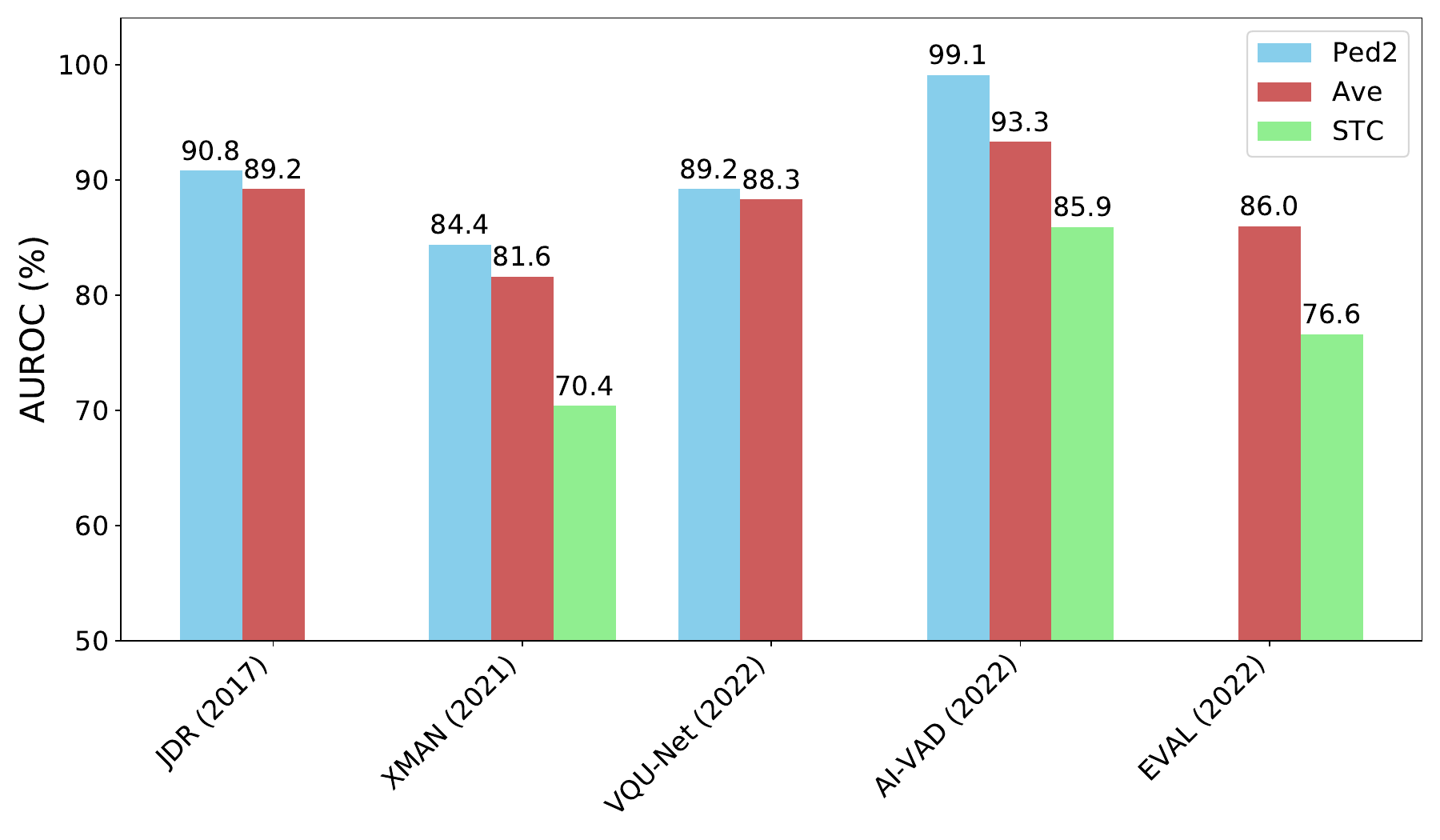}
    \caption{Video anomaly detection performance of Intrinsically interpretable methods on Ped2, Ave, and STC datasets under the semi-supervised setting.}
    \label{fig: vad-performance}
\end{figure}
Intrinsically interpretable methods are the dominant explainable way in video anomaly detection, with strong empirical performance on benchmark datasets as illustrated in Fig.~\ref{fig: vad-performance}. They can be further classified into two categories: a) post-model interpretable methods, and b) in-model attributes interpretation methods.
Post-model interpretable methods explain anomalies by analyzing differences between the actual values and the predictions of the model. These differences reveal normal patterns that the model learns and the abnormal events that diverge from normal videos. In fact, most reconstruction-based or prediction-based methods have explainability to some degree and belong to post-model interpretable methods. This procedure can be formulated as:
\begin{equation}
    \phi(x) = \theta(f(x)-x),
\end{equation}
where $\phi(\cdot)$ shows the heat map of each frame, and $f(\cdot)$ denotes the prediction frame and $x$ is the ground truth frame, and $\theta(\cdot)$ represents the projection of difference of prediction and groundtruth.
For example,~\cite{szymanowicz2022discrete} trained a neural network for predicting future frames. The predicted frame is compared to the ground truth frame and results in a pixel-level error map. Notably, although the error map looks like the saliency map or attention map, they are essentially different. The attention map reveals the weight of each pixel, while the error map indicates the prediction error based on each pixel. Based on the error map, it is easy to locate the error region. The object classification and action recognition models are applied to classify and explain the anomalies.


On the other hand, attribute interpretation methods make use of object attributes to determine whether an object is normal or abnormal. For example,~\cite{singh2022eval} proposed a framework that extracts a high-level depiction of each video segment utilizing networks for appearance and motion. They choose the motion angle and velocity as the attributes to help explain the model. The method for selecting exemplars is applied to choose a representative subset of video segments for a specified spatial region, whose object and its corresponding angle and velocity are kept in the memory bank. During testing, a video sequence is classified as an anomaly if it is different from exemplars. Thus, the explanation can be made by the high-level representation. In particular, the appearance and motion parts of the combined feature vector are mapped to high-level attributes using the last fully connected layer of their respective networks. 
There are some other papers using different attributes.~\cite{hinami2017joint} considered object class and corresponding features and actions. They judge anomaly by predicting anomaly scores of each visual concept. The reason the anomaly can be inferred from these concepts is if one concept is largely deviating from normal patterns.~\cite{reiss2022attribute} regard velocity, pose, and deep as attributes to interpret anomaly. Similarly, these features provide predictions and explanations for anomalies.~\cite{szymanowicz2021x} paid more attention to interactions, so they encoded object category, interaction, and size of bounding box and trained a Gaussian Mixture Model (GMM) from these representations.

In summary, intrinsically interpretable methods can give straightforward explanations without many complicated operations. Although they cannot give some high-level ideas like reasoning-based methods, object attributes perspective is able to explain most anomalies. However, it is important to choose features manually, which requires the human understanding of a particular dataset.

\subsubsection{\bf Memory-based methods}
Another relevant category, often overlapping with reconstruction or prediction-based approaches, is memory-based methods~\cite{park2020learning, yang2022dynamic}. These methods explicitly store representations of normal patterns (e.g., feature vectors, prototypes) in a memory module during training. During inference, test samples are compared against the stored memory items. Anomalies are detected based on their dissimilarity to the closest memory items or their inability to be well-reconstructed using a combination of memory items. Explainability can arise from identifying which memory items are closest to a normal sample versus an anomalous one, or visualizing the reconstruction process using memory items. For instance, if an anomalous sample cannot be well represented by any stored normal pattern, this provides an implicit explanation for its abnormality.


\subsection{Other Related Methods}
Beyond the main categories discussed, several other lines of research contribute to or overlap with explainable visual anomaly detection, although explainability might not be their primary focus. These include feature-based methods, self-supervised approaches, Shapley values, rule-based explanations, and novel class discovery.

\paragraph{\bf Feature-based Methods}
Many modern anomaly detection methods operate in the feature space of pre-trained networks (often trained on large datasets like ImageNet~\cite{deng2009imagenet}). Anomalies are detected based on deviations in these feature representations. While not always explicitly designed for explainability, the features themselves can offer some level of interpretation. These methods often fall into categories like:
\begin{itemize}
    \item \textbf{Feature Reconstruction} These methods train models (like autoencoders) to reconstruct normal features. Anomalies are detected by high reconstruction errors in the feature space~\cite{you2022unified}.
    \item \textbf{Feature Distillation (Student-Teacher)} A "teacher" network (pre-trained on normal data or a larger dataset) guides a "student" network. Anomalies cause discrepancies between student and teacher features~\cite{deng2022reverse, wang2021student}.
    \item \textbf{Feature Comparison/Matching} These methods compare features of test samples to a memory bank of normal features or learn discriminative boundaries in the feature space~\cite{roth2022towards}.
    \item \textbf{Feature Discrimination} Approaches like SimpleNet~\cite{liu2023simplenet} use simple architectures on pre-trained features to discriminate between normal and anomalous patterns effectively.
\end{itemize}
Explainability in these methods often comes from analyzing which features contribute most to the anomaly score or visualizing the differences between anomalous features and the distribution of normal features. However, relating deep features directly back to interpretable concepts in the input space remains a challenge.

\paragraph{\bf Self-Supervised Learning (SSL) Methods}
Self-Supervised Learning (SSL) has become a dominant paradigm in anomaly detection. Methods like CutPaste~\cite{li2021cutpaste} and those using synthetic anomalies~\cite{schluter2022natural} learn representations by solving pretext tasks on normal data. The assumption is that models trained on normal data transformations will struggle to generalize to anomalous data. Explainability here is indirect; the model's failure on the pretext task for an anomalous input serves as the explanation. However, understanding \textit{why} a specific anomaly causes the model to fail on the pretext task requires further investigation, as discussed in Sec.~\ref{sec: 7-future direction}.

\paragraph{\bf Shapley Value Methods}
Originating from cooperative game theory, Shapley values offer a principled way to attribute the prediction outcome to individual input features (e.g., pixels or superpixels). They have been applied to explain anomalies detected by models like autoencoders~\cite{marafon2021explaining} and PCA~\cite{gupta2019shapley}, often using approximations like SHAP (SHapley Additive exPlanations)~\cite{lundberg2017unified, choo2021utilizing}. These methods provide pixel-level importance scores, offering fine-grained explanations but can be computationally expensive.

\paragraph{\bf Rule-Based Explanations}
Another direction aims to interpret unsupervised anomaly detection models by extracting explicit, rule-based explanations~\cite{anonymous2024dissect}. This involves dissecting the internal workings of potentially black-box models to generate human-understandable rules that characterize why certain inputs are flagged as anomalous.

\paragraph{\bf Novel Anomaly Class Discovery}
Explainability is also crucial in scenarios where the goal is not just to detect anomalies but also to discover and characterize potentially novel classes of anomalies, particularly in industrial settings~\cite{anonymous2025anomalyncd}. Understanding the characteristics of newly discovered anomaly types is essential for process improvement and adaptation.

These related areas highlight the diverse strategies employed in visual anomaly detection and the varying degrees and types of explainability they offer.
\section{Differences and Commonalities}
\label{sec: 4-differences-and-commonalities}

\subsection{Explainable methods predominantly applied to images}
From a feasibility perspective, many image anomaly detection methods can be technically applied to video data frame-by-frame. However, considering the intrinsic characteristics of video anomalies (often involving motion and temporal context), several IAD approaches are less suitable or directly applicable to standard VAD benchmarks.

Input-perturbation-based IAD approaches designed for OOD detection or semantic anomaly detection using pretrained image classification models~\cite{liang2018enhancing,Hsu_2020_CVPR,lee2018simple} struggle with VAD. While they can operate frame-wise, image classification networks are typically insensitive to the subtle temporal variations or motion patterns that define many video anomalies (e.g., unusual gaits, interactions). Applying them to object tracks within videos might capture object-level anomalies but would miss motion-based or event-based anomalies, which rely on inter-frame relationships. The explanations derived from input perturbations (sensitivity of classification score to input changes) are less meaningful when the underlying classifier cannot distinguish normal vs. abnormal temporal patterns.

Similarly, certain generative model approaches focused on static image likelihoods, like those learning density functions for typical image classification datasets~\cite{ren2019likelihood,xiao2020likelihood}, are ill-suited for VAD. They primarily capture the distribution of static appearances and fail to model temporal dynamics or complex event structures. Explanations based on likelihood scores from such models would not capture the essence of temporal anomalies.

In summary, explainable IAD methods heavily reliant on static image features or pre-trained image classifiers often lack the necessary mechanisms to model and explain temporal dynamics crucial for VAD. Adapting them effectively requires significant modifications to incorporate temporal context, moving beyond simple frame-wise application.

\subsection{Explainable methods predominantly applied to videos}
Video anomaly detection often focuses on unusual actions or events unfolding over time. Consequently, explainable VAD methods frequently leverage video-specific cues like temporal context, motion information (e.g., optical flow), and object interactions, making them less directly applicable to static images.

Model-specific explanations derived from architectures designed for temporal modeling (e.g., RNNs, 3D CNNs, Transformers processing video clips) inherently rely on temporal data. For instance, methods explaining predictions based on temporal attention weights or gradients through time cannot be applied to single images.

Reasoning-based methods often build representations like scene graphs or causal graphs that explicitly model temporal relationships and interactions between objects over time~\cite{chen2018scene, wu2021learning, doshi2023towards, lv2024video}. While scene graphs can be constructed for static images, they lack the temporal dimension crucial for explaining event-based anomalies (e.g., explaining \textit{why} a sequence of actions is abnormal). The richness of context required for meaningful reasoning-based explanations is often more readily available in video sequences than in isolated images. For example,~\cite{doshi2023towards} constructs a scene graph capturing global interactions and uses metric learning on (subject, predicate, object) triplets derived from video to identify anomalous interactions, providing interpretable feedback on \textit{which} interaction deviates from learned norms.

Methods utilizing optical flow~\cite{Liu_2021_HFVAD} explicitly model motion. Explanations derived from optical flow analysis (e.g., identifying regions with anomalous motion patterns) are inherently video-specific.

Intrinsic interpretable methods using attributes like velocity, pose trajectory, or specific action labels~\cite{hinami2017joint, szymanowicz2021x, singh2022eval, reiss2022attribute} rely on temporal information to define and measure these attributes. While some attributes (like object class) exist in images, many key attributes for explaining video anomalies (e.g., running speed, interaction type) do not.

In essence, explainable VAD methods often exploit the temporal dimension and dynamic content unique to videos, making their direct application or the relevance of their explanations limited in the context of static images.

\subsection{Commonalities and Transferable Approaches}
Despite the differences, certain explainability paradigms can bridge the gap between IAD and VAD, often requiring adaptation.

\begin{figure*}[h]
    \centering
    \subfloat[\small IAD]{\includegraphics[width=0.28\textwidth]{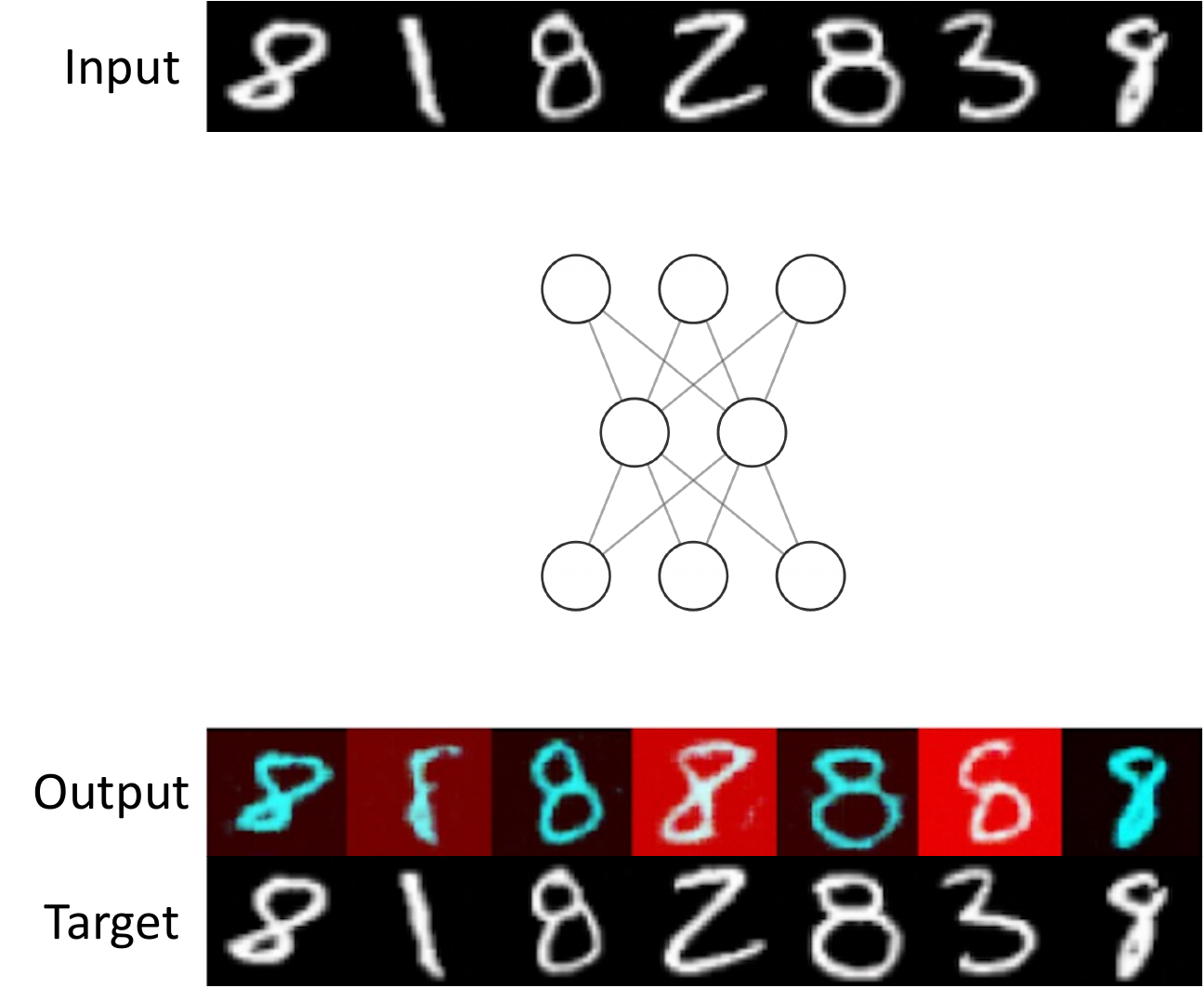}\label{fig: commons-image}}
    \hfill
    \subfloat[\small VAD]{\includegraphics[width=0.7\textwidth]{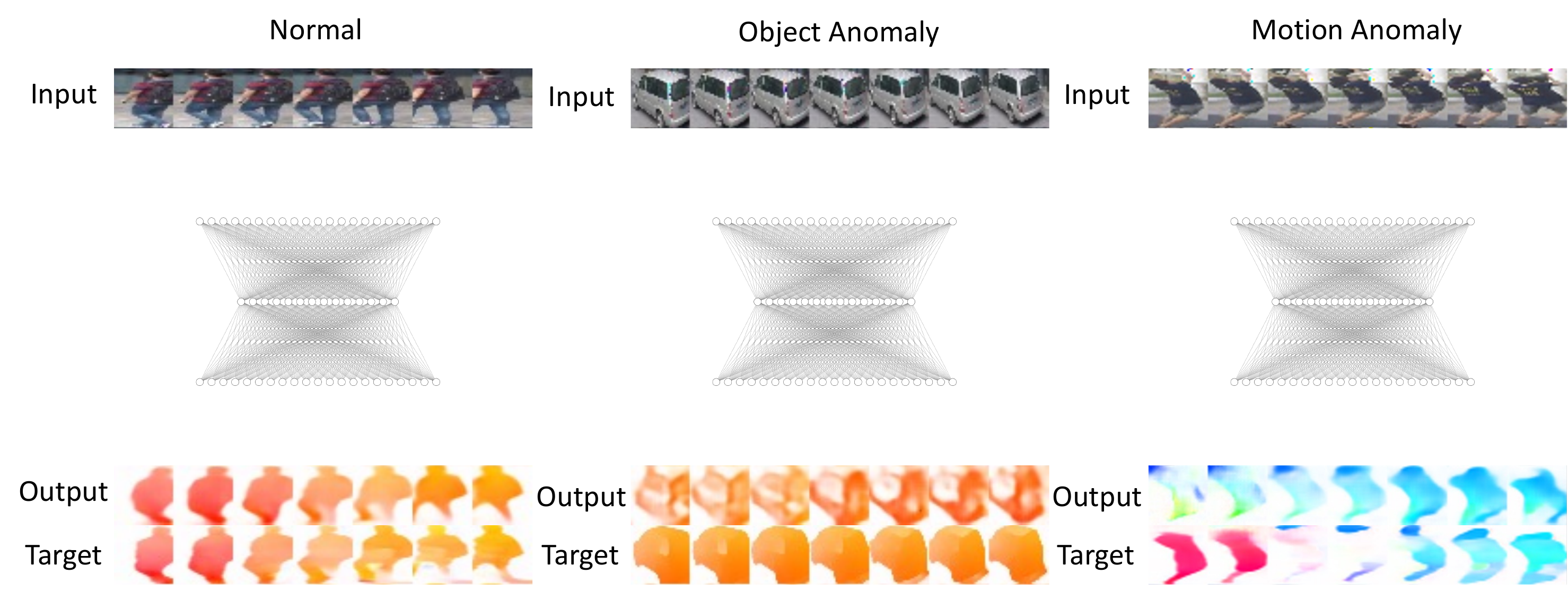}\label{fig: commons-video}}
    \captionsetup{justification=centering} 
    \caption{Error-based anomaly detection methods work for both image and video data.(a) Schematic diagram of error-based image anomaly detection example adapted from~\cite{perera2019ocgan}. The error is the distance between the input image and the reconstruction and is minimized during training. The training data are all number 8s, which are the normal samples. In the inference face, the error will be small for 8s but large for other numbers. (b) Schematic diagram of error-based video anomaly detection example with a window length of $7$ adapted from~\cite{wang2022mrmgavad}. The consecutive frames can be concatenated in channel dimension and then sent into the network for reconstruction. The example here shows the motion branch approach~\cite{wang2022mrmgavad}, where the input is raw frames, and the target is the corresponding optical flow maps. The outputs of the network are required to be close to the target optical flows during training. During the training, all samples are normal data containing normal behavior of pedestrians like walking. During the testing, when encountering normal samples (first column: a walking person), the total error can be minimal since each frame's reconstruction closely approximates the target images. For object anomaly (second column: an unseen object like a car), the reconstruction error of every frame will be large, which shares the same mechanism as error-based image anomaly detection, leading large total error. For motion anomaly (third column:  a person catching a dropping bag while looking upwards), the reconstruction error of certain few frames can be small (last three here), but due to the abruptly changing motion, there must be some frames with extraordinarily large error (first four here), leading to a relatively large error in total. }
    \label{fig: commons}
\end{figure*}

\textbf{Error/Reconstruction-based Methods} Many generative models used for IAD (AEs, VAEs, GANs, Flows, Diffusion Models) rely on reconstruction error or likelihood deviation as the anomaly score~\cite{schlegl2017unsupervised, abati2019latent, liu2020towards, wyatt2022anoddpm, gudovskiy2022cflow}. The resulting error maps (difference between input and reconstruction) serve as a basic form of explanation by highlighting \textit{where} the model failed to reconstruct accurately. This principle can be extended to videos, often by treating video clips as sequences of images or concatenating frames channel-wise (Fig.~\ref{fig: commons}). The rationale is that both object anomalies (consistently high error across frames) and motion anomalies (high error in frames depicting abnormal temporal changes) will result in larger overall reconstruction errors compared to normal sequences. While adaptable, explaining \textit{why} the reconstruction failed (e.g., due to unseen object shape vs. unexpected motion) might require further analysis beyond the raw error map.

\textbf{Attention Mechanisms} Attention maps, whether post-hoc (like Grad-CAM~\cite{8237336}) or learned~\cite{ristea2022self}, highlight salient regions contributing to the model's decision. This concept is applicable to both IAD~\cite{venkataramanan2020attention, liznerski2021explainable} and VAD~\cite{pang2020self}. For videos, attention can be computed frame-wise or across spatio-temporal dimensions. The resulting attention maps offer spatial localization explanations. However, as noted~\cite{cao2022random}, attention might simply highlight foreground objects rather than the anomaly itself, limiting its explanatory power. Evaluating the \textit{faithfulness} of attention maps (i.e., whether they truly reflect the model's reasoning) is crucial~\cite{jacovi2023comprehensive}.

\textbf{Foundation Models (VLMs/LLMs)} Approaches using large pre-trained models like CLIP~\cite{radford2021learning} or GPT-4V~\cite{yang2023dawn} offer promising transferability~\cite{chen2023clip,zhou2023anomalyclip, cao_segment_2023, gu2023anomalyagpt, cao2023towards, zhang2023exploring}. By projecting visual features into a shared space with text embeddings, these models can leverage natural language for explaining anomalies. The same fundamental architecture (visual encoder + language model) can process images or adapted video representations (e.g., using frame pooling~\cite{maaz2023video} or temporal modeling modules~\cite{lv2024video, wang2023vaquita}). The explanation is generated via prompting the language model (e.g., "Describe the anomaly in this image/video."). The key challenge lies in ensuring the generated textual explanations are accurate, relevant, and truly reflect the visual anomaly, rather than being plausible but incorrect hallucinations.

\textbf{Shapley Values / Feature Attribution} Techniques like SHAP~\cite{lundberg2017unified, marafon2021explaining, gupta2019shapley, choo2021utilizing}, which attribute prediction changes to input features, are theoretically applicable to both images (attributing to pixels/patches) and videos (attributing to spatio-temporal regions). They provide fine-grained, mathematically grounded explanations. However, their computational cost can be prohibitive for high-dimensional video data, and interpreting pixel-level attributions for complex temporal events can still be challenging.

\textbf{Challenges in Explainable Visual Anomaly Detection}
Across both modalities, key challenges remain. Defining and \textit{quantifying} the quality of explanations is difficult. Standard AD metrics (AUC, IoU) measure detection/localization performance, not explanatory power. Evaluating aspects like \textit{faithfulness} (does the explanation accurately reflect the model's reasoning?), \textit{stability} (does the explanation change significantly with small input perturbations?), and \textit{effectiveness} (does the explanation help the user understand or debug the model?) requires specialized metrics and user studies~\cite{jacovi2023comprehensive, molnar2019machine}. Furthermore, relating low-level explanations (e.g., attention maps, pixel attributions) to high-level semantic concepts understandable by humans is often non-trivial. Bridging this gap, perhaps by combining different explanation techniques or leveraging foundation models, is an important direction.


\section{Datasets}\label{sec: datasets}
\label{sec: 5-datasets}

In this section, we summarize commonly used public datasets for 2D visual anomaly detection. Understanding the characteristics of these datasets is crucial for evaluating anomaly detection performance and the effectiveness of explainability methods, particularly how well explanations capture the nature of anomalies present in different domains.

\subsection{Image Anomaly Detection Datasets}
A variety of datasets are used for IAD, ranging from standard image classification benchmarks adapted for OOD/semantic anomaly detection to specialized datasets focusing on industrial defects or medical anomalies.

\paragraph{General/OOD Datasets}
For OOD detection and semantic anomaly detection (where one or more classes are treated as normal and others as anomalous), standard image classification datasets are often repurposed. Examples include CIFAR-10 (C10)~\cite{krizhevsky2009learning}, CIFAR-100 (C100)~\cite{krizhevsky2009learning}, MNIST (MN)~\cite{lecun1998gradient}, Fashion-MNIST (fMN)~\cite{xiao2017fashion}, ImageNet~\cite{deng2009imagenet}, Texture~\cite{kylberg2011kylberg}, LSUN~\cite{yu2015lsun}, Caltech-101 (Cal)~\cite{fei2004learning}, CURE-TSR~\cite{temel2017cure}, iSUN~\cite{xu2015turkergaze}, SVHN~\cite{netzer2011reading}, and CelebA~\cite{liu2015deep}. Explainable methods applied here often focus on sample-level discriminativity (e.g., input perturbation, likelihood scores) to explain why an image is classified as out-of-distribution or belonging to an unseen class. The nature of the anomaly is primarily semantic or a domain shift.

\paragraph{Industrial Defect Datasets}
These datasets simulate real-world industrial inspection scenarios and are crucial for evaluating methods aimed at localizing and explaining physical defects. Pixel-level ground truth is often available, facilitating the evaluation of localization-based explanations (e.g., attention maps, reconstruction error maps).
\begin{itemize}
    \item \textbf{MVTec AD (MVTAD)~\cite{bergmann2019mvtec,bergmann2021mvtec}} A widely used benchmark with 5,354 high-resolution images across 15 texture and object categories. Features 73 types of defects (structural changes, missing parts, contamination). Pixel-precise annotations enable evaluation of detection and segmentation/localization explanations.
    \item \textbf{VisA~\cite{zou2022spot}} 10,821 images (9,621 normal, 1,200 anomalous) across 12 object types (e.g., PCBs, capsules). Includes surface and structural defects with image and pixel-level annotations.
    \item \textbf{MVTec LOCO AD (MVTLC)~\cite{bergmann2022beyond}} Focuses on localization, distinguishing between \textit{structural} anomalies (physical defects) and \textit{logical} anomalies (rule violations, e.g., misplaced/missing objects). Contains 3,644 images across 5 categories. Evaluating explanations for logical anomalies poses unique challenges.
    \item \textbf{KSDD2~\cite{bovzivc2021mixed}} Over 3,000 images with fine-grained segmentation masks for surface defects on production items.
    \item \textbf{Magnetic Tile Defect (MTD)~\cite{huang2020surface}} Focuses on real-time defect detection on magnetic tiles.
    \item \textbf{SDD~\cite{tabernik2020segmentation}} Designed for segmenting surface defects, particularly with limited defective samples.
    \item \textbf{BeanTech Anomaly Detection (BTAD)~\cite{mishra2021vt}} 2,830 images of 3 industrial products with pixel-wise anomaly annotations.
    \item \textbf{DAGM~\cite{wieler2007weakly}} From a 2007 competition, focuses on weakly supervised defect detection.
    \item \textbf{MPDD~\cite{jezek2021deep}} Metal parts with defects under various manufacturing conditions.
\end{itemize}
Explaining anomalies in these datasets typically involves localizing the defect (e.g., via heatmaps) and potentially classifying the defect type. Logical anomalies in MVTLC require explanations that go beyond local appearance, potentially involving reasoning about expected object configurations.

\paragraph{Street Scene / Autonomous Driving Datasets}
These datasets focus on identifying unexpected objects or hazards in driving scenarios.
\begin{itemize}
    \item \textbf{Fishyscapes~\cite{blum2019fishyscapes}} Based on Cityscapes~\cite{cordts2016cityscapes}, evaluates pixel-level uncertainty for detecting unusual objects (OOD objects overlaid or from a road hazard dataset). Explainability relates to how well uncertainty estimates correspond to semantic anomalies.
    \item \textbf{RoadAnomaly21~\cite{chan2021segmentmeifyoucan}} Focuses on anomaly segmentation with 100 street scene images with pixel-level annotations for anomalous objects (e.g., unexpected animals, vehicles). 
\end{itemize}
Explanations often involve segmenting the anomalous object and identifying it as out-of-place in the driving context.

\paragraph{Medical Imaging Datasets}
Anomaly detection in medical images is critical, and explainability is paramount for clinical trust.
\begin{itemize}
    \item \textbf{LAG~\cite{li2019attention}} 5,824 fundus images for glaucoma detection (anomaly = glaucoma presence).
    \item \textbf{UK Biobank (UKB)~\cite{sudlow2015uk}:} Large-scale resource including various imaging modalities; can be used for diverse anomaly detection tasks.
    \item \textbf{MSLUB~\cite{lesjak2018novel}} MRI images with multiple sclerosis lesion segmentations.
    \item \textbf{Brain Tumor Segmentation (BRATS)~\cite{menze2014multimodal}:} MRI scans for glioma tumor segmentation.
    \item \textbf{BrainMRI~\cite{kanade2015brain}:} Magnetic resonance images for brain tumor detection, including various tumor types and normal brain scans, processed to evaluate automated detection methods.
    \item \textbf{White Matter Hyperintensities (WMH)~\cite{kuijf2019standardized}} Brain MRI data for segmenting white matter hyperintensities.
    \item \textbf{CheXpert~\cite{irvin2019chexpert}} Large collection of chest radiographs labeled for 14 observations.
    \item \textbf{MedMNIST~\cite{medmnistv1}} Collection of 10 pre-processed medical image datasets for various classification tasks.
\end{itemize}
Explanations typically involve localizing the pathological region (lesion, tumor) and often relating it to specific medical conditions. The need for high fidelity and clinical relevance makes explainability particularly challenging and important.

A summary of representative IAD datasets highlighting their relevance to explainability is presented in Tab.~\ref{tab: iad-datasets}.

\begin{table*}[tbp]
	\caption{Summary of representative Image Anomaly Detection (IAD) datasets relevant to explainability. GT = Ground Truth.}
	\label{tab: iad-datasets}
        \centering
	\setlength\tabcolsep{2.5pt}
        \scalebox{0.85}{
	\begin{tabular}{@{}llcccl@{}}
		\toprule
		\textbf{Dataset} & \textbf{Domain} & \textbf{\# Images} & \textbf{Anomaly Type} & \textbf{Pixel GT} & \textbf{Explainability Relevance} \\ \midrule
		MVTec AD~\cite{bergmann2019mvtec} & Industrial & 5,354 & Structural Defects & Yes & Localization heatmaps, defect classification \\
		VisA~\cite{zou2022spot} & Industrial & 10,821 & Surface/Structural Defects & Yes & Localization heatmaps, defect classification \\
		MVTec LOCO AD~\cite{bergmann2022beyond} & Industrial & 3,644 & Structural \& Logical Defects & Yes & Localization, explaining rule violations (logical) \\
		Fishyscapes~\cite{blum2019fishyscapes} & Driving & (Based on Cityscapes) & OOD Objects & Yes & Uncertainty maps as explanation, semantic segmentation \\
		RoadAnomaly21~\cite{chan2021segmentmeifyoucan} & Driving & 100 (+10) & OOD Objects & Yes & Anomaly segmentation maps \\
		BRATS~\cite{menze2014multimodal} & Medical (MRI) & Varies & Brain Tumors & Yes & Lesion segmentation, clinical relevance \\
		CheXpert~\cite{irvin2019chexpert} & Medical (X-ray) & 224,316 & Pathological Observations & No (Image-level labels) & Explaining image-level diagnosis (e.g., via attention) \\
		CIFAR-10/100~\cite{krizhevsky2009learning} & General & 60k/60k & Semantic/OOD & No & Sample-level explanations (perturbation, likelihood) \\
		\bottomrule
	\end{tabular}
	}
\end{table*}

\subsection{Video Anomaly Detection Datasets}
Video datasets present unique challenges due to the temporal dimension, involving anomalies in actions, interactions, or events unfolding over time.

\begin{table*}[tbp]
	\caption{Comparisons of different Video Anomaly Detection datasets. The UMN dataset does not have official training and testing splits. The UBnormal dataset includes a validation set, which is not displayed here. For setting, ``Se" denotes Semi-supervised, ``We" denotes Weakly-supervised and ``Su" denotes Supervised.}
	\label{tab: vad-datasets}
        \centering
	\setlength\tabcolsep{2.5pt}
        \scalebox{0.9}{
	\begin{tabular}{@{}lcccrrrcr@{}}
		\toprule
		\multicolumn{1}{c}{\multirow{2}{*}{Dataset}} & \multirow{2}{*}{Year} & \multirow{2}{*}{Setting} & \multirow{2}{*}{\#Scenes} & \multicolumn{3}{c}{\# Frames}    & \multirow{2}{*}{\begin{tabular}[c]{@{}c@{}}\# Anomaly\\classes\end{tabular}} & \multicolumn{1}{c}{\multirow{2}{*}{Anomaly classes}}   \\ \cmidrule(lr){5-7}
		\multicolumn{1}{c}{}&  & & & Total& Training& Testing& &                                                                            \\ \midrule
		\multirow{2}{*}{Subway Entrance~\cite{adam2008Subway}}                         & \multirow{2}{*}{2008}     &    \multirow{2}{*}{Se}       & \multirow{2}{*}{1}  & \multirow{2}{*}{144,250}    & \multirow{2}{*}{76,453}    & \multirow{2}{*}{67,797}  & \multirow{2}{*}{5} & \multirow{2}{*}{\begin{tabular}[c]{@{}r@{}} Loitering, Wrong direction, Illegal exit,\\ Unattended objects, Crowding \end{tabular}} \\ 
        &  &    &    &  &  &  & & \\ \hline
		Subway Exit~\cite{adam2008Subway}                               & 2008    &    Se     & 1      & 64,901    & 22,500     & 42,401  & 3 &  Wrong direction, Illegal exit, Unattended objects  \\ \hline
		UMN~\cite{raghavendra2006unusual}	                                 & 2009     &  Se    & 3       & 7,741     & -    	 & -  	   & 1  &  Escaping panic    			                                                            \\ \hline
            \multirow{2}{*}{USCD Ped 1~\cite{2010Ped}}      & \multirow{2}{*}{2010}    & \multirow{2}{*}{Se}  & \multirow{2}{*}{1}  & \multirow{2}{*}{14,000}     & \multirow{2}{*}{6,800}     & \multirow{2}{*}{7,200}   & \multirow{2}{*}{5}  & \multirow{2}{*}{\begin{tabular}[c]{@{}r@{}} Biking, Skating, Small carts, People in wheelchair, \\ People walking across a walkway or in the grass that surrounds it \end{tabular}} \\
            &  &    &    &  &  &  & & \\ \hline
            \multirow{2}{*}{USCD Ped 2~\cite{2010Ped}}      & \multirow{2}{*}{2010}    & \multirow{2}{*}{Se}  & \multirow{2}{*}{1}  & \multirow{2}{*}{4,560}     & \multirow{2}{*}{2,550}     & \multirow{2}{*}{2,010}   & \multirow{2}{*}{5}  & \multirow{2}{*}{\begin{tabular}[c]{@{}r@{}} Biking, Skating, Small carts, People in wheelchair, \\ People walking across a walkway or in the grass that surrounds it \end{tabular}} \\
            &  &    &    &  &  &  & & \\ \hline
		CUHK Avenue~\cite{lu2013Ave}                               & 2013    & Se    & 1         & 30,652    & 15,328    & 15,324  & 5  & Running, Loitering, Throwing objects, Wrong direction, Gathering                                                                   \\ \hline
		\multirow{3}{*}{ShanghaiTech~\cite{zhang2016SHTECH}}                                & \multirow{3}{*}{2017}    & \multirow{3}{*}{Se}    & \multirow{3}{*}{13}   & \multirow{3}{*}{317,398}   & \multirow{3}{*}{274,515}   & \multirow{3}{*}{42,883}  & \multirow{3}{*}{11} & \multirow{3}{*}{\begin{tabular}[c]{@{}r@{}}Biking, Running, Loitering, Vehicle, \\ Skateboarding, Climbing, Throwing objects, Fighting, \\ Jaywalking, Gathering, Abnormal Object \end{tabular}}                                                                   \\ 
            &  &    &    &  &  &  & & \\
            &  &    &    &  &  &  & & \\ \hline
            \multirow{3}{*}{UCF-Crime~\cite{sultani2018UCF}} & \multirow{3}{*}{2018} & \multirow{3}{*}{We} &  \multirow{3}{*}{-}  &  \multirow{3}{*}{13,741,393}  & \multirow{3}{*}{12,631,211} & \multirow{3}{*}{1,110,182} & \multirow{3}{*}{13} &  \multirow{3}{*}{\begin{tabular}[c]{@{}r@{}} Abuse, Arrest, Arson, Assault,  Road Accident,\\ Burglary, Explosion, Fighting, Robbery, \\ Shooting, Stealing, Shoplifting, Vandalism \end{tabular} } \\ 
            &  &    &    &  &  &  & & \\
            &  &    &    &  &  &  & & \\ \hline
            XD Violence~\cite{wu2020XD}                              & 2020   &   We      &   -   &  18,748,800  & -   & - & 6 & Abuse, Car accident, Explosion, Fighting, Riot, Shooting   \\ \hline
		\multirow{7}{*}{Street Scene~\cite{ramachandra2020street}}                              & \multirow{7}{*}{2020}   &    \multirow{7}{*}{Se}     & \multirow{7}{*}{1}      & \multirow{7}{*}{203,257}   & \multirow{7}{*}{56,847}    & \multirow{7}{*}{146,410} & \multirow{7}{*}{17} &  \multirow{7}{*}{\begin{tabular}[c]{@{}r@{}} Jaywalking, Biker outside lane, \\ Loitering,  Dog on sidewalk, Car outside lane, \\ Worker in bushes, Biker on sidewalk,  Pedestrian reverses direction, \\ Car u-turn, Car illegally parked, Person opening trunk, \\ Person exits car on street, Skateboarder in bike lane, \\ Person sitting on bench,  Metermaid ticketing car, \\ Car turning from parking space, Motorcycle drives onto sidewalk \end{tabular}}       \\ 
            &  &    &    &  &  &  & & \\
            &  &    &    &  &  &  & & \\
            &  &    &    &  &  &  & & \\
            &  &    &    &  &  &  & & \\
            &  &    &    &  &  &  & & \\
            &  &    &    &  &  &  & & \\ \hline
		\multirow{2}{*}{TAD~\cite{lv2021localizing}}                               & \multirow{2}{*}{2021}                  &  \multirow{2}{*}{We}  &  \multirow{2}{*}{-}  & \multirow{2}{*}{540,272} & \multirow{2}{*}{452,220} &  \multirow{2}{*}{88,052}  & \multirow{2}{*}{7}  &  \multirow{2}{*}{ \begin{tabular}[c]{@{}r@{}} Vehicle accidents, Illegal turns, Illegal occupations, \\ Retrograde motion, Pedestrian on road, Road spills, The else \end{tabular}}      \\ 
            &  &    &    &  &  &  & & \\ \hline
		\multirow{6}{*}{UBnormal~\cite{acsintoae2022ubnormal}}       & \multirow{6}{*}{2022}  & \multirow{6}{*}{Su}     & \multirow{6}{*}{29}         & \multirow{6}{*}{236,902}   & \multirow{6}{*}{116,087}   & \multirow{6}{*}{92,640}  & \multirow{6}{*}{22}   &  \multirow{6}{*}{\begin{tabular}[c]{@{}r@{}} Running, Falling, Fighting, Sleeping, \\ Crawling, Having a seizure, Laying down, Dancing,\\ Stealing, Rotating $360$ degrees, Shuffling, Walking injured,\\ Walking drunk, Stumbling walk, People and car accident, \\ Car crash, Running injured, Fire, Smoke, \\ Jaywalking, Driving outside lane, Jumping \end{tabular} }   \\ 
            &  &    &    &  &  &  & & \\
            &  &    &    &  &  &  & & \\
            &  &    &    &  &  &  & & \\
            &  &    &    &  &  &  & & \\
            &  &    &    &  &  &  & & \\ \hline
		\multirow{8}{*}{NWPU Campus~\cite{cao2023new}}                                & \multirow{8}{*}{2023}  &  \multirow{8}{*}{Se}  & \multirow{8}{*}{43}          & \multirow{8}{*}{1,466,073} & \multirow{8}{*}{1,082,014} & \multirow{8}{*}{384,059} & \multirow{8}{*}{28}  &  \multirow{8}{*}{\begin{tabular}[c]{@{}r@{}} Climbing fence, Car crossing square, Cycling on footpath, \\ Kicking trash can,  Jaywalking, Snatching bag, Crossing lawn, \\ Wrong turn, Cycling on square, Chasing, Loitering, \\ Scuffle, Littering, Forgetting backpack, U-turn, \\ Battering,  Driving on wrong side,  Falling, \\ Suddenly stopping cycling in the middle of the road,  Group conflict, \\  Climbing tree, Stealing, Illegal parking, Trucks,  Protest,\\ Playing with water, Photographing in restricted area, Dogs\end{tabular} } \\
            &  &    &    &  &  &  & & \\
            &  &    &    &  &  &  & & \\
            &  &    &    &  &  &  & & \\
            &  &    &    &  &  &  & & \\
            &  &    &    &  &  &  & & \\
            &  &    &    &  &  &  & & \\
            &  &    &    &  &  &  & & \\
         \bottomrule
	\end{tabular}}
\end{table*}

The datasets commonly used for VAD are summarized in Tab.~\ref{tab: vad-datasets}, which includes details on settings, frame counts, anomaly classes, and relevance to explainability evaluation. Below, we briefly describe some key datasets and their implications for explainable VAD.

\paragraph{Early Benchmarks (UMN~\cite{raghavendra2006unusual}, Ped1/Ped2~\cite{2010Ped}, Subway~\cite{adam2008Subway}, Avenue~\cite{lu2013Ave})}
These datasets feature relatively simple scenes and anomalies (e.g., running crowds, vehicles in pedestrian zones, wrong direction). They are useful for evaluating basic spatio-temporal localization explanations (e.g., attention maps, reconstruction errors). Ped1 provides pixel masks for some clips, allowing direct evaluation of localization accuracy. However, the limited complexity might not challenge more sophisticated reasoning or attribute-based explanation methods.

\paragraph{Complex Scene Benchmarks (ShanghaiTech~\cite{zhang2016SHTECH}, Street Scene~\cite{ramachandra2020street}, NWPU Campus~\cite{cao2023new})}
These datasets offer more diverse scenes, camera angles, lighting conditions, and a wider range of anomaly types, including complex interactions and subtle events. They are better suited for evaluating advanced explanation methods:
\begin{itemize}
    \item \textbf{STC} Its diversity supports reasoning-based explanations (e.g., scene graphs capturing interactions). The X-MAN re-annotation~\cite{szymanowicz2021x} provides explicit reason labels, directly facilitating evaluation of attribute-based explanations.
    \item \textbf{Street Scene} Features numerous, varied, naturally occurring anomalies. Good for evaluating fine-grained localization and explanations related to specific actions or interactions.
    \item \textbf{NWPU Campus} Very large-scale with many scenes and anomaly types, including scene-dependent anomalies (e.g., fighting is normal in a boxing ring, abnormal elsewhere). This explicitly challenges methods to provide context-aware explanations, likely requiring reasoning capabilities.
\end{itemize}

\paragraph{Weakly-Supervised Benchmarks (UCF-Crime~\cite{sultani2018UCF}, XD-Violence~\cite{wu2020XD}, TAD~\cite{lv2021localizing})}
These datasets typically provide only video-level labels (indicating if an anomaly occurs in a long video, often without precise timing or location). Explaining anomalies here is challenging, often requiring methods that can link coarse labels to specific spatio-temporal segments (e.g., Multiple Instance Learning approaches). Generating fine-grained explanations like pixel maps or detailed attributes is difficult due to the lack of precise ground truth.
\begin{itemize}
    \item \textbf{UCF-Crime/XD-Violence} Focus on real-world criminal/violent events.
    \item \textbf{TAD} Focuses on traffic anomalies, providing object-level labels for evaluating explanations in this specific domain.
\end{itemize}

\paragraph{Specialized/Supervised Benchmarks (X-MAN~\cite{szymanowicz2021x}, UBnormal~\cite{acsintoae2022ubnormal})}
These datasets are specifically designed with explainability or detailed supervision in mind.
\begin{itemize}
    \item \textbf{X-MAN} As mentioned, provides multiple textual reason labels per anomalous frame (action, object, location), directly targeting the evaluation of attribute-based explanation methods.
    \item \textbf{UBnormal} Provides pixel-level annotations in a supervised setting (anomalies seen during training, but test set is open-set). Useful for evaluating precise localization explanations and supervised attribute learning in a controlled (virtual) environment.
\end{itemize}

Overall, the choice of dataset significantly influences the type of explainable method that can be developed and evaluated. Datasets with fine-grained annotations (pixel masks, bounding boxes, attribute labels) are essential for quantitatively evaluating localization and attribute-based explanations. More complex datasets with diverse scenes and context-dependent anomalies are needed to push research towards more robust reasoning-based explanations.

\section{Evaluation Metrics}\label{sec: evaluation-metrics}
\label{sec: 6-metrics}

Evaluating explainable anomaly detection involves assessing both the accuracy of the anomaly detection itself and the quality of the generated explanations. Standard metrics focus on detection/localization performance, while evaluating explanations requires considering aspects like faithfulness, stability, and utility.

\subsection{Metrics for Anomaly Detection Performance}
These metrics quantify how well the system identifies and localizes anomalies, irrespective of the explanation provided.
\paragraph{Image-level / Frame-level Detection}
For tasks determining whether an entire image or video frame is anomalous, standard binary classification metrics are used:
\begin{itemize}
    \item \textbf{Area Under the Receiver Operating Characteristic (AUROC)} Plots True Positive Rate (TPR) vs. False Positive Rate (FPR) across different thresholds. A score of 1 indicates perfect separation, 0.5 indicates random guessing.
    \item \textbf{Area Under the Precision-Recall curve (AUPR)} Plots Precision vs. Recall. More informative than AUROC for highly imbalanced datasets typical in anomaly detection.
    \item \textbf{F1-Score} The harmonic mean of Precision and Recall at a specific threshold.
\end{itemize}

\paragraph{Pixel-level / Region-level Localization}
For tasks requiring localization of anomalous regions within images or video frames:
\begin{itemize}
    \item \textbf{Pixel-wise AUROC/AUPR} Calculated by treating each pixel as an individual classification instance. Can be biased towards larger anomalous regions~\cite{bergmann2019mvtec}.
    \item \textbf{Intersection over Union (IoU)} Measures the overlap between the predicted anomalous region $P$ and the ground truth anomalous region $G$ at a given threshold $t$: $ \text{IoU} = |P \cap G| / |P \cup G| = \text{TP} / (\text{TP} + \text{FP} + \text{FN}) $.
    \item \textbf{Per-Region Overlap (PRO)~\cite{bergmann2021mvtec}} Averages the detection rate over each ground truth connected component, giving equal weight to small and large anomalies. $ \text{PRO} = \frac{1}{N} \sum_i \sum_j \frac{|P_i \cap G_{i,j}|}{|G_{i,j}|} $, where $N$ is the total number of GT components, $P_i$ is the prediction for image $i$, and $G_{i,j}$ is the $j$-th ground truth component in image $i$.
    \item \textbf{Track-Based Detection Rate (TBDR)~\cite{ramachandra2020street} (Video)} Measures the fraction of anomalous object tracks correctly detected (a track is detected if $\alpha$ fraction of its GT regions are detected with IoU $\geq \beta$). Used with False Positive Rate per frame (FPR).
    \item \textbf{Region-Based Detection Rate (RBDR)~\cite{ramachandra2020street} (Video):} Measures the fraction of all anomalous regions across all frames correctly detected (IoU $\geq \beta$). Also evaluated against FPR.
\end{itemize}

\subsection{Metrics for Evaluating Explanations}
Evaluating the quality of explanations themselves is more challenging and less standardized. Key aspects include~\cite{jacovi2023comprehensive, molnar2019machine}:
\begin{itemize}
    \item \textbf{Faithfulness (Fidelity)} Does the explanation accurately reflect the model's internal reasoning process? For feature attribution methods (like attention or SHAP), one common approach is to measure the change in model output when perturbing or removing features deemed important by the explanation. A faithful explanation should lead to a significant drop in the anomaly score when important features are removed.
    \item \textbf{Plausibility (Sanity)} Does the explanation make sense to a human observer? This is often evaluated qualitatively through user inspection. For example, does an attention map highlight the visually anomalous region? Does a textual explanation correctly describe the anomaly? Sanity checks can also involve testing whether explanations change appropriately when model parameters or inputs are randomized~\cite{adebayo2018sanity}.
    \item \textbf{Stability (Robustness)} Is the explanation robust to small, insignificant perturbations in the input? An unstable explanation that changes drastically with minor input variations may not be reliable.
    \item \textbf{Completeness:} Does the explanation capture all the factors influencing the decision?
    \item \textbf{Usefulness/Effectiveness} Does the explanation help the user achieve a specific goal, such as debugging the model, understanding its limitations, or making a more informed decision? This often requires task-specific user studies.
    \item \textbf{Consistency with Ground Truth Explanations} For datasets like X-MAN~\cite{szymanowicz2021x} that provide ground truth reasons for anomalies (e.g., attribute labels), the consistency between the model's explanation (e.g., predicted attributes) and the ground truth can be directly measured using metrics like accuracy or F1-score for attribute prediction.
\end{itemize}
Currently, quantitative evaluation of explainability in visual anomaly detection often relies on proxy tasks. For localization-based explanations (attention maps, error maps), the overlap with ground truth anomaly masks (using IoU or pixel AUROC) is frequently used as a measure of explanation quality, conflating localization performance with explanation faithfulness. For attribute-based explanations, performance on predicting ground truth attributes (if available) serves as a proxy. Developing more direct and standardized metrics for faithfulness, stability, and usefulness remains an open research area, especially for complex visual anomalies. Recent work has also focused on developing practical metrics specifically for explaining abnormal uncertainty in related tasks like object detection~\cite{giloni2025dil}, which might offer insights for evaluating uncertainty-based explanations in anomaly detection.


\section{Future Directions and Open Problems}
\label{sec: 7-future direction}
Based on the reviewed literature and remaining challenges, we outline several promising future directions and open problems in explainable 2D visual anomaly detection.

\subsection{Quantifying and Evaluating Explanation Quality}
\textit{Challenge:} While numerous methods provide visual or textual explanations, quantitatively evaluating their quality beyond standard detection/localization metrics (Sec.~\ref{sec: 6-metrics}) remains a major hurdle. Current evaluations often rely on qualitative examples or proxy metrics (e.g., localization accuracy for attention maps) that don't fully capture explanation fidelity or usefulness.
\textit{Directions:}
\begin{itemize}
    \item \textbf{Developing Faithfulness Metrics} Adapt and standardize metrics from general XAI research~\cite{jacovi2023comprehensive} to assess if explanations (e.g., heatmaps from attention or Shapley values~\cite{marafon2021explaining, gupta2019shapley}) truly reflect the model's decision process for visual anomalies. This involves measuring performance degradation upon perturbing explained features/regions.
    \item \textbf{Metrics for Textual Explanations} For explanations generated by VLMs/LLMs~\cite{gu2023anomalyagpt, cao2023towards}, develop metrics beyond semantic similarity to evaluate factual correctness, relevance to the visual anomaly, and absence of hallucination.
    \item \textbf{User Studies} Conduct targeted user studies to evaluate the \textit{effectiveness} and \textit{usefulness} of different explanation types (e.g., heatmaps vs. attribute lists vs. natural language) for specific tasks (e.g., debugging industrial defects, understanding medical anomalies).\cite{molnar2019machine}
    \item \textbf{Benchmarking Explanation Stability} Develop protocols to measure the robustness of explanations to minor input variations, ensuring reliability.
    \item \textbf{Ground Truth Explanations} Encourage the creation of more datasets with ground truth explanations (like X-MAN~\cite{szymanowicz2021x}) to enable direct evaluation of attribute-based or reason-based explanations.
\end{itemize}

\subsection{Explaining Diverse Anomaly Detection Paradigms}
\textit{Challenge:} Explainability needs to be integrated into and tailored for various AD settings and techniques.
\textit{Directions:}
\begin{itemize}
    \item \textbf{Self-Supervised Learning (SSL)} Current SSL-based AD methods~\cite{tack2020csi,Georgescu2021ssmtl,wang2022self, li2021cutpaste, schluter2022natural} often lack clear explanations for \textit{why} an anomaly causes the pretext task to fail. Future work should investigate methods to link pretext task performance degradation back to specific anomalous features or temporal inconsistencies.
    \item \textbf{Few-Shot, Zero-Shot, and Open-Set AD} Explaining decisions in data-scarce settings (few-shot~\cite{lu2020SAAD}, zero-shot~\cite{deng2023anovl}) or when encountering completely novel anomalies (open-set) is crucial. How can models justify flagging something as anomalous when they have limited or no prior exposure? Explanations might involve highlighting deviation from known normality or leveraging external knowledge (e.g., via VLMs).
    \item \textbf{Weakly Supervised AD} In settings with only video-level labels~\cite{sultani2018UCF, wu2020XD, feng2021mist}, developing methods that not only localize anomalies but also provide faithful explanations for the localization remains challenging.
    \item \textbf{Feature-Based Methods} For methods operating on pre-trained features~\cite{you2022unified, deng2022reverse, wang2021student, roth2022towards, liu2023simplenet}, develop techniques to translate feature-level anomaly explanations (e.g., deviations in feature space) into interpretable concepts in the input space.
\end{itemize}

\subsection{Context-Aware and Reasoning-Based Explanations}
\textit{Challenge:} Many anomalies are context-dependent (e.g., a person running is normal on a track but abnormal in a library). Current explanations often focus on local features and struggle with contextual reasoning.
\textit{Directions:}
\begin{itemize}
    \item \textbf{Scene-Context Modeling} Develop explainable models that explicitly incorporate scene context (e.g., using scene graphs~\cite{doshi2023towards}, knowledge graphs~\cite{nesen2020knowledge}, or attention mechanisms that consider background) to justify why an object or event is anomalous \textit{in that specific context}. The NWPU Campus dataset~\cite{cao2023new} is a good testbed for this.
    \item \textbf{Causal Explanations} Move beyond correlational explanations (e.g., attention maps) towards causal explanations~\cite{wu2021learning} that identify the underlying causes of anomalies, especially for complex events in videos.
    \item \textbf{Integrating Commonsense Knowledge} Leverage external commonsense knowledge bases or the implicit knowledge in foundation models to provide more human-like reasoning for why certain situations are anomalous.
\end{itemize}

\subsection{Explainable Foundation Models for Visual Anomaly Detection}
\textit{Challenge:} While foundation models (VLMs/LLMs)~\cite{achiam2023gpt, liu2023llava, kirillov2023SAM, Liu2023DINO} show promise for zero-shot AD and generating textual explanations~\cite{gu2023anomalyagpt, cao2023towards, zhang2023exploring}, ensuring the faithfulness and controllability of these explanations is crucial. Applying them effectively to VAD also requires addressing temporal modeling.
\textit{Directions:}
\begin{itemize}
    \item \textbf{Faithful Textual Explanations} Develop methods to ensure that the natural language explanations generated by LLMs accurately reflect the visual evidence and the model's internal anomaly detection process, mitigating hallucination.
    \item \textbf{Interactive Explanations} Explore interactive explanation systems where users can query the model (e.g., "Why is this region anomalous?", "What would need to change for this to be normal?") to gain deeper understanding.
    \item \textbf{Temporal Reasoning with Foundation Models} Adapt foundation models for robust temporal reasoning in VAD, enabling explanations that capture complex event dynamics, object interactions, and motion anomalies~\cite{maaz2023video, lv2024video, wang2023vaquita}.
    \item \textbf{Efficiency and Controllability} Improve the efficiency of using large foundation models for AD and develop ways to control the level of detail and perspective of the generated explanations.
\end{itemize}

\subsection{Real-World Considerations: Efficiency and Robustness}
\textit{Challenge:} Practical deployment requires explainable AD systems to be efficient and robust.
\textit{Directions:}
\begin{itemize}
    \item \textbf{Efficient Explanations} Develop explanation techniques (especially for complex methods like Shapley values or reasoning-based approaches) that are computationally efficient enough for real-time applications.
    \item \textbf{Robustness of Explanations} Evaluate and improve the robustness of explanations against adversarial attacks specifically designed to mislead the explanation module while potentially leaving the anomaly detection itself unaffected.
    \item \textbf{Domain Adaptation and Generalization} Ensure that explainable AD methods generalize well to new domains and that the explanations remain meaningful and consistent across different operational conditions.
\end{itemize}


\section{Conclusion}
\label{sec: 8-conclusion}

This paper presents the first survey focused specifically on explainable anomaly detection for 2D visual data (images and videos). We provided a comprehensive overview of the growing body of work in explainable image anomaly detection (IAD) and explainable video anomaly detection (VAD), categorizing methods based on their core techniques, including attention, input perturbation, generative models, reasoning, intrinsic attributes, and foundation models. We elaborated on how these methods aim to explain the anomaly identification process and analyzed the factors influencing their applicability across image and video modalities. Furthermore, we summarized relevant datasets, highlighting their suitability for different types of anomalies and explanation evaluations, and discussed evaluation metrics, covering both standard performance measures and crucial aspects of explanation quality like faithfulness and utility. The survey concludes by identifying key open challenges and promising future directions.





{
\small
\bibliographystyle{IEEEtran}
\bibliography{tpami}
}

\section{Biography Section}
 
\vspace{11pt}

\begin{IEEEbiography}[{\includegraphics[width=1in,height=1.25in,clip,keepaspectratio]{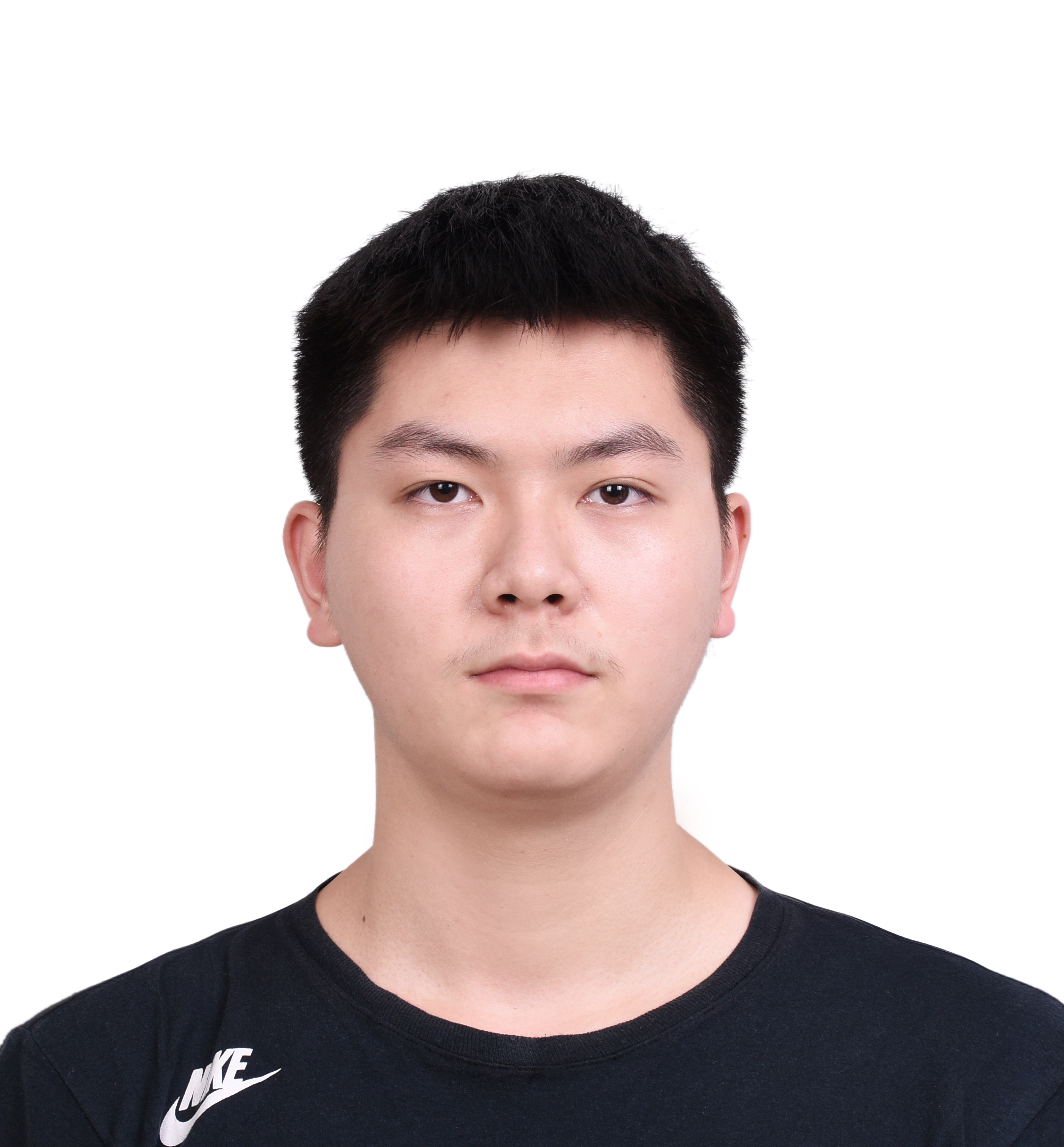}}]{Yizhou Wang} (Student Member, IEEE) received the B.S. degree in Mathematics and Applied Mathematics (Honors Program) from the School of Mathematics and Statistics, Xi'an Jiaotong University, Xi'an, China, in 2020, and the M.S. degree in Electrical and Computer Engineering from Northeastern University, Boston, MA, in 2023. He is currently working toward the Ph.D. degree in the Department of Electrical and Computer Engineering, Northeastern University, Boston, Massachusetts, under the supervision of Prof. Yun Raymond Fu. His research interests include machine learning, computer vision and anomaly detection. He has published some papers at top-tier conferences including ACL, ICLR, CVPR, CIKM, ICDM, IJCAI, WACV, IJCNN and top-tier journals including TKDE, Nature Communications and SIIMS. He has served as a Reviewer for conferences including ICML, NeurIPS, ICLR, CVPR, ECCV, ICCV, KDD, SDM, AAAI, IJCAI, PAKDD, ICME, ACCV and journals including TPAMI, IJCV, TNNLS, PR, TKDD, IoS, KAIS, Journal of Big Data, IJFS, IEEE Trans. Veh. Technol., and CIM.

\end{IEEEbiography}
\begin{IEEEbiography}[{\includegraphics[width=1in,height=1.25in,clip,keepaspectratio]{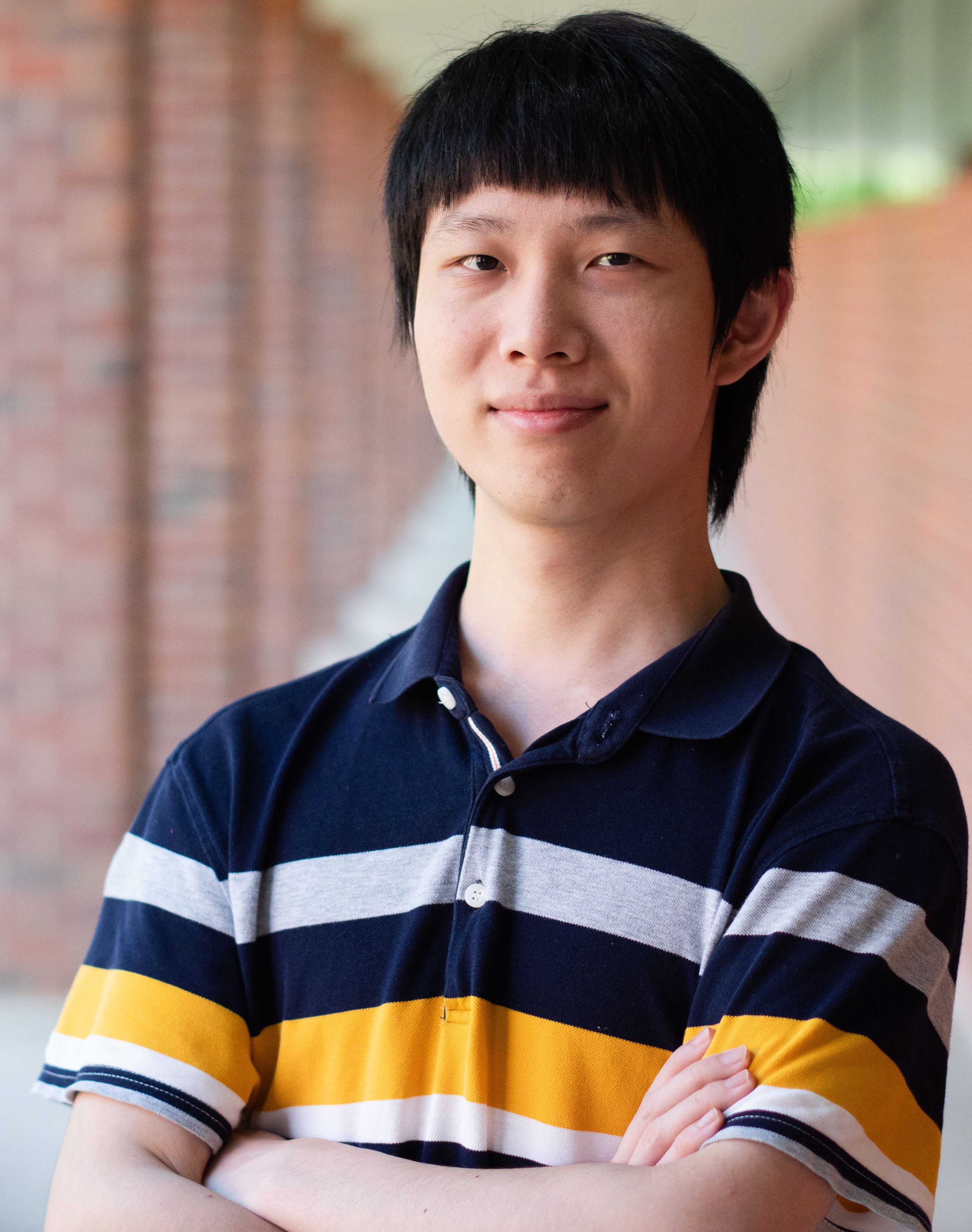}}]{Dongliang Guo}
received the B.Eng. degree in software engineering from University of Electronic Science and Technology of China, Chengdu, China, in 2020. He is presently pursuing his Ph.D. at the University of Virginia, Charlottesville, VA, USA, under the supervision of Prof. Sheng Li. His research interests include trustworthy representation learning and multimodal learning. He has published papers at top-tier conferences, including ICLR, SDM. He is a reviewer for several IEEE Transactions including TKDD, TPAMI, TMLR, and is a program committee member for NIPS, ICLR, ICML, CVPR, KDD, AAAI, SDM and so on.
\end{IEEEbiography}

\begin{IEEEbiography}[{\includegraphics[width=1in,height=1.25in,clip,keepaspectratio]{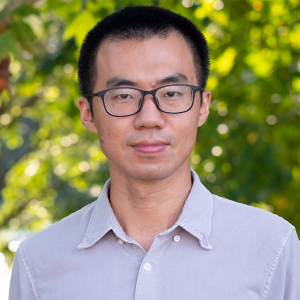}}]{Sheng Li} (S'11-M'17-SM'19) received the B.Eng.~degree in computer science and engineering and the M.Eng.~degree in information security from Nanjing University of Posts and Telecommunications, China, and the Ph.D.~degree in computer engineering from Northeastern University, Boston, MA, in 2010, 2012 and 2017, respectively. He is an Associate Professor at the School of Data Science, University of Virginia. He was a Tenure-Track Assistant Professor at the School of Computing, University of Georgia from 2018 to 2022, and was a data scientist at Adobe Research from 2017 to 2018. He has published over 150 papers at peer-reviewed conferences and journals, and has received over 10 research awards, such as the INNS Aharon Katzir Young Investigator Award, Fred C. Davidson Early Career Scholar Award, Adobe Data Science Research Award, Cisco Faculty Research Award, SDM Best Paper Award, and IEEE FG Best Student Paper Honorable Mention Award. He serves as an Associate Editor of IEEE Transactions on Neural Networks and Learning Systems, IEEE Transactions on Circuits and Systems for Video Technology, and IEEE Computational Intelligence Magazine. He has also served as area chair for IJCAI, NeurIPS, ICML, ICLR, and SDM. His research interests include trustworthy representation learning, computer vision, and causal inference.
\end{IEEEbiography}

\begin{IEEEbiography}[{\includegraphics[width=1in,height=1.25in,clip,keepaspectratio]{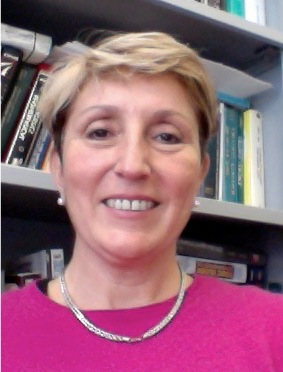}}]{Octavia Camps}
(Member, IEEE) received the B.S. degree in computer science and the B.S. degree in electrical engineering from the Universidad de la Republica, Montevideo, Uruguay, in 1981 and 1984, respectively, and the M.S. and Ph.D. degrees in electrical engineering from the University of Washington, Seattle, WA, USA, in 1987 and 1992, respectively.
Since 2006, she has been a Professor with the Electrical and Computer Engineering Department, Northeastern University, Boston, MA, USA. From 1991 to 2006, she was a faculty of Electrical Engineering and of Computer Science and Engineering, The Pennsylvania State University, State College, PA, USA. She was a visiting Researcher with the Computer Science Department, Boston University, Boston, MA, USA, during Spring 2013 and in 2000, she was a visiting faculty with the California Institute of Technology, Pasadena, CA, USA, and with the University of Southern California, Los Angeles, CA, USA. Her main research interests include dynamics-based computer vision and machine learning.
Dr. Camps is currently an Associate Editor of IEEE Transactions on Pattern Analysis and Machine Intelligence (PAMI).
\end{IEEEbiography}

\begin{IEEEbiography}[{\includegraphics[width=1in,height=1.25in,clip,keepaspectratio]{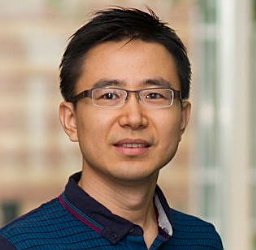}}]{Yun Fu}  (Fellow, IEEE) received the B.Eng. degree
in information engineering and the M.Eng. degree
in pattern recognition and intelligence systems from
Xi'an Jiaotong University, China, and the M.S.
degree in statistics and the Ph.D. degree in electrical
and computer engineering from the University of
Illinois at Urbana–Champaign. He has been an Interdisciplinary Faculty Member with the College of
Engineering and the College of Computer and Information Science, Northeastern University, since 2012
His research interests are machine learning, computational intelligence, big data mining, computer vision, pattern recognition,
and cyber-physical systems. He has extensive publications in leading journals,
books/book chapters, and international conferences/workshops. He serves
as an associate editor, the chair, a PC member, and a reviewer of many
top journals and international conferences/workshops. He received seven
prestigious young investigator awards from NAE, ONR, ARO, IEEE, INNS,
UIUC, and Grainger Foundation; eleven best paper awards from IEEE, ACM,
IAPR, SPIE, and SIAM; and many major industrial research awards from
Google, Samsung, Amazon, Konica Minolta, JP Morgan, Zebra, Adobe, and
Mathworks. He is currently an Associate Editor of the IEEE TRANSACTIONS
ON PATTERN ANALYSIS AND MACHINE INTELLIGENCE. He is Member of Academia Europaea (MAE), Member of European Academy of Sciences and Arts (EASA), Fellow of National Academy of Inventors (NAI), AAAS Fellow, AAAI Fellow, IEEE Fellow, AIMBE Fellow, IAPR Fellow, OSA Fellow, SPIE Fellow, AAIA Fellow, and ACM Distinguished Scientist.
\end{IEEEbiography}


%

\vfill

\end{document}